%% file: main.tex
\documentclass[manuscript]{acmart}
\usepackage{hyperref}
\usepackage{tikz} % I edit figures in IPE and export them in TikZ
\usetikzlibrary{arrows.meta,patterns,ipe}
\tikzset{
every picture/.style={
  execute at end picture={
    \path (current bounding box.south west) +(-5,-5) (current bounding box.north east) +(5,5);
    }
  }
}
\usepackage{url}
\usepackage{enumitem} % to customize enumerate
\usepackage{amsthm} % to get proofs
\usepackage{amssymb} % to get supsetneq
\usepackage{svg} % to get SVG support
\usepackage{algorithm} 
\usepackage{algpseudocode} % to get pseudocode
\algblock[]{At}{End} 
\usepackage{subfiles} % Best loaded last in the preamble
\usepackage{subcaption} % to get subfigure
\usepackage{dsfont} % to get mathds

\newtheorem{theorem}{Theorem}
\newtheorem{remark}{Remark}
\newtheorem{lemma}[theorem]{Lemma}
\newtheorem{definition}{Definition}
\newtheorem{assumption}{Assumption}
\newtheorem{corollary}{Corollary}

\DeclareMathOperator*{\Clim}{C-lim} % define Cesaro limit symbol
\DeclareMathOperator*{\argmin}{argmin} % define argmin limit symbol
\def\MDP{\mathcal{M}}
\def\hatMDP{\widehat{\mathcal{M}}}
\def\roundMDP{\mathring{\mathcal{M}}}
\def\WISC{\texttt{WISC}}

\def\EWISC{\texttt{EWISC}}
\def\mathEWISC{\mathtt{EWISC}}
\def\BLINQ{\texttt{BLINQ}}
\def\mathBLINQ{\mathtt{BLINQ}}
\def\QWhittle{\texttt{Q-Whittle}}
\def\WIQLUCB{\texttt{WIQL-UCB}}
\def\QWI{\texttt{QWI}}
\def\QGI{\texttt{QGI}}
\def\QWINN{\texttt{QWINN}}

\newcommand{\rand}[1]{#1} % notation for random variables
\newcommand{\indicPi}[1]{\mathds{1}_{#1 \in \pi}} % notation for random variables
\newcommand{\calA}{\mathcal{A}}
\newcommand{\calS}{\mathcal{S}}
\newcommand{\norminf}[1]{\left\lVert #1 \right\rVert_\infty}
\newcommand{\normone}[1]{\left\lVert #1 \right\rVert_1}
\newcommand{\range}[1]{\text{sp}(#1)}
\newcommand{\diam}[1]{D(#1)}

\newcommand{\blue}[1]{#1}

\begin{document}

\title{Model-Based Learning of Whittle indices}
\date{November 2025}
\author{Joël Charles-Rebuffé}
\email{joel.charles-rebuffe@univ-grenoble-alpes.fr}
\affiliation{%
  \institution{Univ. Grenoble Alpes, Inria, CNRS, Grenoble INP}
  \city{38000 Grenoble}
  \country{France}
}
\author{Nicolas Gast}
\email{nicolas.gast@inria.fr}
\affiliation{%
  \institution{Univ. Grenoble Alpes, Inria, CNRS, Grenoble INP}
  \city{38000 Grenoble}
  \country{France}
}
\author{Bruno Gaujal}
\email{bruno.gaujal@inria.fr}
\affiliation{%
  \institution{Univ. Grenoble Alpes, Inria, CNRS, Grenoble INP}
  \city{38000 Grenoble}
  \country{France}
}

\begin{abstract}
  We present \BLINQ, a new model-based algorithm that learns the Whittle indices of an indexable, communicating and unichain Markov Decision Process (MDP). Our approach relies on building an empirical estimate of the MDP and then computing its Whittle indices using an extended version of a state-of-the-art existing algorithm. We provide a proof of convergence to the Whittle indices we want to learn as well as a bound on the time needed to learn them with arbitrary precision. Moreover, we investigate its computational complexity. Our numerical experiments suggest that \BLINQ~significantly outperforms existing $Q$-learning approaches in terms of the number of samples needed to get an accurate approximation. In addition, it has a total computational cost even lower than $Q$-learning for any reasonably high number of samples. These observations persist even when the $Q$-learning algorithms are sped up using neural networks to predict Q-values.
\end{abstract}

% ACM keywords

% CCS classification


\maketitle

\section{Introduction}
\label{sec:introduction}
\subfile{sections/introduction.tex}

\section{Definition of Whittle Indices}
\label{sec:model}
\subfile{sections/model.tex}

\section{Topology of indexable MDPs}
\label{sec:topology_I}
\subfile{sections/topology_indexable_arms.tex}

\section{\EWISC: an index computing algorithm for all MDPs}
\label{sec:EWISC}
\subfile{sections/EWISC.tex}

\section{\BLINQ: a new framework for learning Whittle indices}
\label{sec:BLINQ}
\subfile{sections/BLINQ.tex}

\section{Numerical experiments}
\label{sec:numerical}
\subfile{sections/num_exp.tex}

\section{Conclusion}
\label{sec:conclusion}

In this paper we propose a new lerning framework called \BLINQ~to learn Whittle indices. Contrary to most existing approaches that are model-free and use $Q$-learning, \BLINQ~is model-based: it uses a simulator of an unknown MDP whose we want to learn the Whittle indices, and builds an estimate of it. \BLINQ~sporadically calls an algorithm \EWISC~that uses this estimate to approximate the original Whittle indices. Thanks to our algorithm \EWISC, \BLINQ~can learn the Whittle indices even when its estimated MDP is non-indexable. We prove that for a communicating unichain MDP with a Markovian exploration policy, the values computed by \BLINQ~converge to the true indices of the simulated MDP at a rate $O(1/\sqrt{T})$ where $T$ is the number of time steps elapsed since the start of the learning.

For future research, one important questions is the design of an efficient exploration policy (instead of uniform used in \BLINQ). For instance, if \BLINQ~is used in an online framework with multi-arms and where rewards actually matter, it is important to properly balance the exploration of potentially interesting arms and the exploitation of the arms that we believe are good, i.e. minimizing the \emph{regret}. Even in a pure exploration framework, one might want to target specific state-action pair to help us distinguish between various indices. In particular, an algorithm that learns the index based policy should learn correctly the \emph{order} of the indices but not necessarily the exact \emph{values} of the indices. 

Also, Theorem \ref{th:linear_conv}, Lemma \ref{lem:learning_time} and Theorem \ref{th:precision_time} give performance guarantees using constants that depend on the unknown MDP. If we want to explicitely compute a sufficient learning time to get an accurate approximation of the Whittle indices, it would be of great interest to learn these constants, as tightly as possible.

Lastly, yet another research path would be to expand the set of MDPs \BLINQ~works on by using more general Whittle index computing algorithms. For instance, one could consider the multichain case.

% \subsubsection*{Author Contributions}
% If you'd like to, you may include a section for author contributions as is done
% in many journals. This is optional and at the discretion of the authors. Only add
% this information once your submission is accepted and deanonymized. 

\appendix

\subfile{sections/appendix.tex}

\bibliography{main}
\bibliographystyle{unsrt}

\end{document}

%% file: sections/introduction.tex
Restless multi-armed bandits (RMABs) can model several resource allocation problems. To cite only a few applications, RMABs are used to solve optimization problems related to stochastic scheduling in a queue \cite{verloop2016asymptotically,ansell_whittles_2003,nino2024stochastic}, opportunistic scheduling \cite{borkar2017opportunistic,tang2020minimizing,xiong2024whittle}, machine maintenance \cite{glazebrook2005index}, healthcare \cite{ijcai2021p556} or recommendation systems \cite{wang2020restless}. See also \cite{nino2023markovian,gittins2011multi} for surveys on RMABs.

A RMAB is composed of $N$ independent arms, each of these arms being a Markov Decision Process that evolves in discrete time. At each time instant, the decision maker chooses a set of arms to be activated with the goal of maximizing either rewards averaged over an infinite horizon ---\emph{average} reward criterion--- or cumulated discounted rewards ---\emph{discounted} reward criterion---. In the literature, this problem was first investigated from a computational point of view: how to efficiently compute an optimal policy. In a seminal paper~\cite{gittins1979dynamic}, Gittins and Jones introduced the concept of index policies and showed that a policy based on these indices---now known as Gittins index policy---is optimal for some particular RMABs. In \cite{whittle1988restless}, Whittle introduced a generalization of these indices---the Whittle indices---that greatly extends the applicability of index policies. These indices can be computed for all arms that satisfy a condition called \emph{indexability}. While Whittle index policies are not optimal in general, they have been shown to be asymptotically optimal when the number of arms goes to infinity, under indexability and a technical condition that is often satisfied in practice \cite{weiss_weber_1990,gast2023exponential}. Moreover, they work extremly well in practice \cite{nino2023markovian}.

To implement an index policy, one needs to compute, for each arm, an \emph{index function} that maps each of its states to a real value, called the \emph{index} of this state. The index policy consists in activating the arm whose current state has the highest index. The computation of this index function of a given arm can be done independently of the other arms and only requires to have access to the arm parameter, \emph{i.e}, its transition matrices $P$ and reward functions $r$. There exist in the literature very efficient algorithms and implementations that compute the indices of an arm given $(P,r)$ \cite{akbarzadeh2019restless,chakravorty2014multi,gast2023testing}.

However, there are many situations  where the parameters $(P,r)$ are unknown to the decision maker who can only estimate them through sampling. This is the focus of the paper: we consider a scenario where an agent interacts with an environment that can simulate an arm. The agent is faced with the learning problem of how to use these samples to estimate the indices as precisely as possible.

In this paper, we develop a new model-based approach to learn Whittle indices of communicating and unichain MDPs. Our approach consists in sampling a MDP and building estimates of its transition matrices and reward vectors $(\hat{P},\hat{r})$. The algorithm of \cite{gast2023testing} (which we call \WISC~ for \textbf{W}hittle \textbf{I}ndex in \textbf{S}ub\textbf{C}ubic time) is the fastest known algorithm to compute the Whittle indices of an indexable and unichain MDP. Hence, it would be tempting to simply call \WISC~on $(\hat{P},\hat{r})$ and return the indices of these estimates. Yet, we show that this \emph{naive} method does not work in general because the set of indexable MDPs is not an open-set: even if the true MDP with parameters $(P,r)$ is indexable, there is no guarantee that the estimated MDP with parameters $(\hat{P},\hat{r})$ is indexable, regardless of how close $(\hat{P},\hat{r})$ is from $(P,r)$. This is why we introduce an algorithm that generalizes \WISC, called \textbf{E}xtended \WISC~or \EWISC. It takes as input any parameter values $(\hat{P},\hat{r})$ and outputs a vector of real values $\EWISC(\hat{P},\hat{r})$. We show that if the unknown MDP $(P,r)$ is indexable, then $\EWISC(\hat{P},\hat{r})$ converges to the true indices $\WISC(P,r)$ when $(\hat{P},\hat{r})$ converges to $(P,r)$, regardless of the indexability of the parameters $(\hat{P},\hat{r})$. This allows us to derive an end-to-end learning framework that we call \BLINQ, which stands for \textbf{B}LINQ \textbf{L}earning \textbf{I}s \textbf{N}ot \textbf{Q}-learning. We prove that if $(P,r)$ is indexable, then \BLINQ~almost surely learns its true indices as the number of samples goes to infinity with an upper bound on the convergence speed. Our theoretical proof focuses on the average reward criterion but it also works for the discounted average criterion, as we illustrate in our numerical experiments. They show that \BLINQ~largely outperforms $Q$-learning and Neural Network based approaches for any reasonable number of samples.

\blue{
    \paragraph*{Related work} The problem of learning Whittle indices has become popular in the recent years, and a number of reinforcement learning (RL) algorithms have been developed. Their underlying approaches can be classified into four different categories: Neural Networks (NN), $Q-$Learning, a mix of both, and model-based. A representation of existing algorithms for these four types of RL algorithms can be found in Figure (\ref{fig:literature_review}).
    
    For $Q-$learning, one of the first contributions is \cite{borkar2018reinforcement} which focuses on a particular case of monotonic bandit model. It has been extended to (unstructured) discrete state-space in \cite{fu2019towards} which develops a first pseudo-code called \texttt{QWIC}. A two-timescale approach is developed in \cite{robledo_borkar2022deeplearning} with an algorithm called \QWI. This algorithm is adapted for the more restrictive case of Gittins indices in \cite{dhankar2025tabulardeepreinforcementlearning}, giving birth to the \QGI~algorithm. $Q-$learning approaches are modular and lightweight in terms of memory usage and computing power, but they often need a lot of samples to converge and they are very sensitive to the choice of stepsizes: \cite{Kakarapalli2024faster_Q_learning} explores these inherent limitation as well as the question of what exploration policy is best suited. All the papers we cited so far use a model with discounted rewards: the reward at time $t$ is multiplied by $\gamma^t$ for some predefined discount factor $\gamma < 1$. It is important to note that the convergence of $Q-$learning based approaches is generally easier to prove and requires less constraining assumptions with discounted rewards. In contrast, our work focuses on average reward and works well for both criteria, as explained in Section \ref{subsec:discountedEWISC} and illustrated in Section \ref{sec:numerical}. However there also exists $Q-$learning based approaches with the average reward criterion such as \QWhittle~\cite{avrachenkov2022whittle}, which is a small modification of \QWI. Finally, a new algorithm called \WIQLUCB~\cite{jonah2026WIQL} implements some ideas like in \cite{Kakarapalli2024faster_Q_learning} so that it does not need fine-tuning stepsizes.

    As far as NNs are concerned, the main algorithm is \texttt{NeurWIN} from \cite{nakhleh2021neurwin}. It works only for the discounted reward criterion and requires the MDP to be strongly indexable, which is more constraining than the majority of literature regarding Whittle indices ---see footnote at the bottom of Section \ref{par:charact_of_WI}---. It has been tweaked in some more specific models, for instance in \cite{hao2024NeurWINapplied}. However, as discussed in \cite{jonah2026WIQL}, NN-based approaches often require a substantial amount of computing power, and their training phase and high sensitivity to the NN and MDP parameters make their results sometimes difficult to reproduce. In contrast, as we will explain in Section \ref{sec:numerical}, our learning framework is very efficient in terms of computing power and does not require fine-tuned parameters. 

    Some works try to get the best of both worlds by combining $Q-$learning and NNs. A first attempt has been made for the discounted reward criterion in \cite{avrachenkov2022whittle}: the authors improve the performance of \QWI~by creating \QWINN. Later attempts include \texttt{Neural Q-Whittle} \cite{xiong2023finite}, which is based on \QWhittle~and works for the average reward criterion. It gives an upper bound on the convergence time needed to accurately learn the Whittle indices, and is particularly relevant for many-states MDPs thanks to the upscaling ability of NNs. Lastly, the \texttt{DQN} algorithm from \cite{Mittal2024DQN} implements a similar approach as \QWINN. As usual with NNs, it comes at the cost of high computing power usage.

    On the other hand, model-based approaches usually allow for very accurate learning with only few samples needed, at the cost of memory usage: by definition, the decision maker needs to store all estimated parameters of the unknown MDP. A first model-based approach is \texttt{RB-TSDE} \cite{akbarzadeh2023learning} which provides a Bayesian regret analysis. However it suffers from two main difficulties. First, it assumes that the estimated MDP is indexable, given that the unknown MDP is also indexable. We prove in Section \ref{sec:topology_I} that this assumption does not hold in general: there exists non-indexable, arbitrarily close estimates of indexable MDPs. Then, it provides an analysis that relies on a relatively strong ergodicity hypothesis \cite[Assumptions 3 or 5]{akbarzadeh2023learning}. Another model-based approach is \texttt{UC-Whittle} from \cite{wang2023optimistic} but for the discounted reward criterion, with a worst-case regret analysis. It uses a Whittle index-like policy that does not require indexability, making our work the first ---to the best of our knowledge--- to tackle the problem of non-indexable estimates of indexable MDPs.
    \begin{figure}[hbtp]
        \centering
        \begin{subfigure}[t]{\textwidth}
            \centering
            \input{IPE_figures/literature_review.tex}  
        \end{subfigure}
        \caption{An overview of existing algorithms to learn Whittle indices. For each category, algorithms are presented from left to right by chronological order. For each algorithm, we indicate if it is made for discounted of average reward criterion, the main model and learning assumptions it needs, as well as its convergence rate ($\emptyset$ for no convergence guarantee).}  
        \Description{Visual representation of our literature review}
        \label{fig:literature_review}
    \end{figure}
}

\paragraph*{Contributions} We introduce and analyze a new model-based algorithm to learn Whittle index, that incurs four contributions. Our first contribution is to show that for some indexable MDPs there exists arbitrarily close, non indexable MDPs. As a result, we extend the \WISC~algorithm by introducing \EWISC, which works on all MDPs (indexable or not) and returns values that coincide with Whittle indices when the input is indexable. Our second contribution is to show that when Whittle indices of an indexable MDP are all distinct, then it has an indexable neighborhood. Our third and main contribution is the introduction and detailed analysis of \BLINQ: a learning framework that uses \EWISC. We provide a detailed analysis of its performance, that is, its rate of convergence as a function of the time and MDP parameters. As a side product of our analysis, our fourth contribution is to provide an easy-to-compute bound on the Whittle indices of an indexable MDP, in terms of its parameters.

\paragraph*{Roadmap} The rest of the paper is organized as follows. In Section~\ref{sec:model}, we introduce notations, we define indexability as well as the Whittle index policy. In Section~\ref{sec:topology_I}, we study the topogy of indexable MDPs: we show that the set of indexable MDPs is not an open set with Theorem \ref{th:non-indexable_neighborhood}, and that when all Whittle indices of an indexable MDP are distinct then it has an indexable neighborhood thanks to Theorem \ref{th:distinct_WI}. In Section~\ref{sec:EWISC} we introduce \EWISC: an algorithm that takes as input all MDPs ---indexable and non-indexable---and whose output are the Whittle indices when the input is indexable. We then show in Theorem \ref{th:linear_conv} that the output of \EWISC~is Lipschitz continuous in the MDP parameters, regardless of its indexability. In Section \ref{sec:BLINQ}, we introduce \BLINQ: our learning framework to learn Whittle indices. We quantify its speed of convergence to the true Whittle indices using two results. In Lemma \ref{lem:learning_time} we provide a performance guarantee on the time needed for \BLINQ~to accurately learn the parameters of a MDP, and then we use it in Theorem \ref{th:precision_time} to provide a performance guarantee on the time needed to accurately learn the Whittle indices of an indexable MDP. This also provides an explicit bound on the Whittle indices of an indexable MDP, as a side product, in Corollary \ref{cor:WI_bounded}. In Section~\ref{sec:numerical} we present our numerical experiments.

%% file: IPE_figures/literature_review.tex
\begin{tikzpicture}[ipe import]
  \draw
    (16, 640)
     -- (416, 640);
  \node[ipe node, anchor=center]
     at (40, 752) {$Q$-learning};
  \draw
    (16, 736)
     -- (416, 736);
  \draw
    (16, 704)
     -- (416, 704);
  \draw
    (64, 768)
     -- (64, 640);
  \node[ipe node, anchor=center, font=\small]
     at (96, 752) {specific model};
  \node[ipe node, anchor=center, font=\small]
     at (96, 760) {[19]};
  \node[ipe node, anchor=center, font=\small]
     at (156, 752) {discounted};
  \node[ipe node, anchor=center, font=\small]
     at (156, 760) {\texttt{QWIC} [20]};
  \node[ipe node, anchor=center, font=\footnotesize]
     at (156, 744) {$\emptyset$};
  \node[ipe node, anchor=center, font=\small]
     at (280, 752) {discounted};
  \node[ipe node, anchor=center, font=\small]
     at (280, 760) {\texttt{QWI} [21]};
  \node[ipe node, anchor=center, font=\footnotesize]
     at (280, 744) {$o(1)^*$};
  \node[ipe node, anchor=center, font=\small]
     at (220, 752) {average$^{\dagger}$};
  \node[ipe node, anchor=center, font=\small]
     at (220, 760) {\texttt{Q-Whittle} [24]};
  \node[ipe node, anchor=center, font=\footnotesize]
     at (220, 744) {$o(1)^{*}$};
  \node[ipe node, anchor=center, font=\small]
     at (324, 752) {Gittins};
  \node[ipe node, anchor=center, font=\small]
     at (324, 760) {\texttt{QGI} [22]};
  \node[ipe node, anchor=center, font=\footnotesize]
     at (324, 744) {$o(1)^*$};
  \node[ipe node, anchor=center, font=\small]
     at (380, 752) {average$^{\ddagger}$};
  \node[ipe node, anchor=center, font=\small]
     at (380, 760) {\texttt{WIQL-UCB} [25]};
  \node[ipe node, anchor=center, font=\footnotesize]
     at (380, 744) {$o(1)$, stepsizes provided};
  \node[ipe node, anchor=center]
     at (40, 728) {Neural};
  \node[ipe node, anchor=center]
     at (40, 716) {Networks};
  \draw
    (16, 672)
     -- (416, 672);
  \node[ipe node, anchor=center]
     at (40, 696) {$Q$-learning};
  \node[ipe node, anchor=center]
     at (40, 684) {+NNs};
  \node[ipe node, anchor=center]
     at (40, 664) {Model};
  \node[ipe node, anchor=center]
     at (40, 652) {based};
  \node[ipe node, anchor=center, font=\small]
     at (184, 720) {discounted$^\mathsection$};
  \node[ipe node, anchor=center, font=\small]
     at (184, 728) {\texttt{NeurWIN} [26]};
  \node[ipe node, anchor=center, font=\footnotesize]
     at (184, 712) {$o(1)^*$};
  \node[ipe node, anchor=center, font=\small]
     at (260, 728) {\texttt{NeurWIN} tweaked [27]};
  \node[ipe node, anchor=center, font=\small]
     at (184, 688) {discounted};
  \node[ipe node, anchor=center, font=\small]
     at (184, 696) {\texttt{QWINN} [21]};
  \node[ipe node, anchor=center, font=\footnotesize]
     at (184, 680) {$\emptyset$};
  \node[ipe node, anchor=center, font=\small]
     at (260, 688) {average$^\dagger$};
  \node[ipe node, anchor=center, font=\small]
     at (260, 696) {\texttt{Neural-Q-Whittle} [28]};
  \node[ipe node, anchor=center, font=\footnotesize]
     at (260, 680) {$o(1 / T^{2 / 3})^*$};
  \node[ipe node, anchor=center, font=\small]
     at (336, 688) {discounted};
  \node[ipe node, anchor=center, font=\small]
     at (336, 696) {\texttt{DQN} [29]};
  \node[ipe node, anchor=center, font=\footnotesize]
     at (336, 680) {$\emptyset$};
  \node[ipe node, anchor=center, font=\small]
     at (136, 656) {average$^{\# \dagger}$};
  \node[ipe node, anchor=center, font=\small]
     at (136, 664) {\texttt{RB-TSDE} [30]};
  \node[ipe node, anchor=center, font=\footnotesize]
     at (136, 648) {Regret analysis};
  \node[ipe node, anchor=center, font=\small]
     at (240, 656) {discounted$^\dagger$};
  \node[ipe node, anchor=center, font=\small]
     at (240, 664) {\texttt{UC-Whittle} [31]};
  \node[ipe node, anchor=center, font=\footnotesize]
     at (240, 648) {Regret analysis of Whittle-like policy};
  \node[ipe node, anchor=center, font=\small]
     at (356, 656) {average$^\dagger$ or discounted};
  \node[ipe node, anchor=center, font=\small]
     at (356, 664) {\texttt{BLINQ} (this work)};
  \node[ipe node, anchor=center, font=\footnotesize]
     at (356, 648) {Index learning in $O(1 / \sqrt{T})$};
  \draw
    (16, 768)
     -- (416, 768);
  \node[ipe node, anchor=north west, font=\footnotesize]
     at (16, 636) {$^*$Convergence depends on stepsizes and/or NN parameters};
  \node[ipe node, anchor=north west, font=\footnotesize]
     at (328, 636) {$^{\mathsection}$strongly indexable};
  \node[ipe node, anchor=north west, font=\footnotesize]
     at (192, 636) {$^{\#}$strong ergodicity+indexability of estimate};
  \node[ipe node, anchor=north west, font=\footnotesize]
     at (16, 628) {$^{\dagger}$unichain$_{}$};
  \node[ipe node, anchor=north west, font=\footnotesize]
     at (52, 628) {$^{\ddagger}$communicating};
\end{tikzpicture}

%% file: sections/model.tex
\subsection{The N-arm bandit model}

An $N$-arm Markovian bandit problem is a discrete-time Markov decision processes (MDP) composed of $N$ sub-components. Each of these components is called an \emph{arm} and is a two-action MDP.  At each time step, the decision maker can activate up to $M$ arms for some predefined $M \leq N$. Subsequently, each arm produces a reward and the states of all arms evolve independently according to their Markovian kernel. The decision maker is interested in maximizing the long-term average reward, summed over all arms.

As computing an optimal policy for such a problem is computationally difficult \cite{papadimitriou1999_PSPACE}, we focus on a particular policy called the \emph{Whittle index policy} \cite{whittle1988restless}. To each state of each arm we associate a real number called \emph{Whittle index} representing \emph{how good activating the arm in that state is} (see \S\ref{ssec:Whittle} for a precise definition of these indices). Following that, the Whittle index policy simply consists in activating the $M$ arms that have the highest Whittle indices.  This policy is known to be asymptotically optimal when the number of arms grows large and under some other technical assumptions \cite{weiss_weber_1990}. Because the Whittle indices are computed independently on each arm, in the rest of this paper we focus on a single arm and design an algorithm that learns all its Whittle indices. 

\blue{We emphasize that our learning framework is model-based: it learns the parameters of an unknown arm and computes its indices. However in the rest of the paper we only focus on learning the transition probabilities and not the rewards.} This is because the main difficulty does not lie in learning the rewards: this can add complexities to our equations and proofs but not conceptual issues to our problem. As a result we choose to consider MDPs with known and deterministic rewards, as most learning papers do \cite{avrachenkov2022whittle,dhankar2025tabulardeepreinforcementlearning}.

% {\color{red} Joël si on justifie l'emploi de politiques stationnaires, voit peut-être Puterman sec 4.4, sec discounted 6.2.1}

\subsection{Definition of an arm}
\label{subsec:def_MDP}

An arm $\MDP$ is a Markov decision process (MDP) with a finite state space $\calS := \{ 0, \dots, S - 1 \}$ and an action space $\calA := \{ 0, 1 \}$. Action $0$ is called \emph{passive} and action $1$ is called \emph{active}. For each $a \in \calA$, the reward vector of $\MDP$ under action $a$ is $(r^a_s)_{s \in \calS}$, and its transition kernel is $\big(P^a_{s,s'}\big)_{s,s' \in \calS}$. If action $a \in \calA$ is taken while the arm is in state $s \in \calS$, its next state is $s' \in \calS$ with probability $P^a_{s, s^{'}}$.  This generates a reward $r^a_s$. The matrices $P^a$ for $a \in \calA$ are of dimension $S \times S$. 

\blue{A deterministic policy $\pi$ is a choice of action for each state. In the following, $\pi$ denotes the set of states in which action $1$ is chosen: $\pi \subseteq \calS$ where $s \in \pi$ if and only if (iff) $\pi$ chooses action $1$ in state $s$. Let us designate by $\indicPi{s}$ the binary number equal to $1$ when $s \in \pi$ and $0$ otherwise. The transition matrix of the Markov chain induced by policy $\pi$ is then simply $P^\pi := ( P^{\indicPi{s}}_{s, s'} )_{s, s' \in \calS}$, and the reward vector induced by policy $\pi$ is $r^\pi := (r^{\indicPi{s}}_s)_{s \in \calS}$.}

\blue{In the rest of the paper, we focus on unichain MDPs, i.e. where every policy incurs a single recurrent set of states.}

\subsection{Bellman-optimal policies and activation advantage}
\label{subsec:BO_and_advantage}

In this paper, we seek for policies that maximize the long-term average reward, that is often called the gain. Following \cite{puterman2014markov}, for a policy $\pi \subseteq \calS$, we denote by $g^\pi_s$ the gain of $\pi$ when starting in state $s$ as 
\begin{equation*}
    \forall s \in \calS, g^\pi_s := \Clim_{t \rightarrow \infty} \mathbb{E}\left[ r^{\indicPi{\rand{S_t}}}
    _{\rand{S_t}} \right],
\end{equation*}
where $(\rand{S_t})_t$ is  a  Markov chain with transition matrix $P^\pi$ starting at $\rand{S_0} := s$ a.s.\ and $\Clim$ stands for  the Cesàro limit: $\Clim_{t \rightarrow \infty} u_{t} := \lim{T' \rightarrow \infty}\frac{1}{T'}\sum_{t' = 0}^{T'} u_{t'}$. By \cite[Chapter 8]{puterman2014markov}, there is also a bias vector $(b^\pi_s)_{s \in \calS}$ such that for every state $s$:
\begin{equation}
    g_s^\pi + b_s^\pi = r^\pi_s + \sum_{s' \in \calS} P^{\indicPi{s}}_{s,s'} b^\pi_{s'}.\label{eq:def_bias}
\end{equation}
For any finite-state finite-action MDP, the gain is well defined \cite{puterman2014markov}. Moreover, when the MDP is unichain, $g^\pi_s$ does not depend on $s$ and the bias is defined up to an additive constant \cite{puterman2014markov}, i.e. if two vectors $b,b'$ satisfy (\ref{eq:def_bias}) then $b-b' \in \mathbb{R}e$. As the MDPs we study are unichain, we define the \emph{activation advantage of state $s$ under policy $\pi$:}
\begin{equation*}
\label{def:act_advantage}
\begin{split}
    \alpha_s^\pi & := r^1_s + \sum_{s' \in \calS} P^1_{s,s'} b^\pi_{s'} - \left(r^0_s + \sum_{s' \in \calS} P^0_{s,s'} b^\pi_{s'}\right) = r^1_s - r^0_s + \big( P^1_{s,\cdot} - P^0_{s,\cdot} \big) \cdot b^\pi,
\end{split}
\end{equation*}
which does not depend on the choice of the vector $b^\pi$ that satisfies Equation (\ref{eq:def_bias}). $P^a_{s,\cdot}$ designates the row vector $\big(P^a_{s,s'}\big)_{s' \in \calS}$, and $b^\pi$ is considered as a column vector. The activation advantage can be thought as a measurement of \emph{how optimal action $1$ is} (if $\alpha_s^\pi>0$) or \emph{how sub-optimal} it is (if $\alpha_s^\pi<0$), compared to action $0$.

We say that a policy $\pi^*$ is \emph{gain optimal} if for all state $s$ and all policy $\pi$, $g^{\pi^*}_s\ge g^\pi_s$. It is known that such a policy exists \cite{puterman2014markov}. In practice, there might exist numerous gain optimal policies. Hence, to define Whittle index, the authors of \cite{gast2023testing} use a refined notion of optimality, called Bellman optimality\footnote{This notion is also known as canonical optimality, see \cite{gast2023optimal,yushkevich1974class}.}:
\begin{definition}\label{def:BO}
A policy $\pi^*$ is \emph{Bellman optimal (BO)} if there exists a vector $b \in \mathbb{R}^S$ such that for all $s\in\calS$:
\begin{equation}
    g^{*} + b_s = \max_{a \in \calA} \left\{ r^a_s + \sum_{s' \in S} P^a_{s,s'} b_{s'} \right\}. \label{eq:BO}
\end{equation}
\end{definition}
Note that $b$ is a bias of policy $\pi^*$. A Bellman optimal policy is also gain optimal, but the converse is not true \cite{gast2023testing}. 

The activation advantage is a very handy tool for characterizing BO policies, see \cite[Lemma 2.(i)]{gast2023testing}:
\begin{equation}
    \label{eq:charact_BO}
    \pi \text{ is BO} \Longleftrightarrow \forall s \in \pi, \alpha_s^\pi \geq 0 \text{ and } \forall s \in \calS \setminus \pi, \alpha_s^\pi \leq 0.
\end{equation}
Because of that, it is at the heart of both the \WISC~algorithm of \cite{gast2023testing} and our generalization \EWISC, see Algorithm \ref{alg:EWISC}.

\subsection{Definition of Whittle index}
\label{ssec:Whittle}

For a MDP $\MDP=(\calS, \calA, r, P)$, let us take a real-valued parameter $\lambda \in \mathbb{R}$ and define the \emph{$\lambda$-penalized MDP $\MDP (\lambda)$} by substracting $\lambda$ to all active rewards, i.e. the reward vector $(r^a_s)_{a \in \calA, s \in \calS}$ is replaced by $r(\lambda)$: for each $s \in \calS$ and $a \in \{0,1\}, {r}(\lambda)^a_s := r^a_s - \lambda a$. The quantity $\lambda$ is a \emph{penalty for activation}. The Whittle indices of all states are defined using penalities as follows.
\begin{definition}
    Let $\MDP$ be an MDP. We say that $\MDP$ is indexable if for any penalties $\lambda, \lambda' \in \mathbb{R}$ that satisfy $\lambda < \lambda'$, and for any BO policies $\pi, \pi'$ in the $\lambda$-penalized MDP $\MDP(\lambda)$ and $\lambda'$-penalized MDP $\MDP(\lambda')$, respectively, we have: $\pi' \subseteq \pi$.
    For an indexable MDP, the Whittle index of the state $s$ is the maximal penalty $\lambda_s$ such that there exists a BO policy that contains $s$. That is, $\forall \lambda < \lambda_s, \text{ all BO policies in } \MDP(\lambda) \text{ contain } s$, and $\forall \lambda > \lambda_s, \text{ no BO policy in } \MDP(\lambda) \text{ contain } s$.
\end{definition}

The condition of indexability can be thought as a ``monotonicity'' condition: the set of states in which action $1$ is chosen by a BO policy in the $\lambda-$penalized MDP $\MDP(\lambda)$ decreases when the penalty grows. \blue{It is required to define properly Whittle indices. Indexability also implies some facts such as \cite[Lemma 2.(iii)]{gast2023exponential} that are needed for proving the asymptotic optimality of the Whittle index policy.}

\paragraph{Alternative characterizations of Whittle indices}
\label{par:charact_of_WI}

There exists multiple characterizations of Whittle indices. Here we recall the one of \cite[Lemma 1]{gast2023testing}, which we use to construct our algorithm \EWISC.
\begin{lemma}
    \label{lem:characterization_WI}
    Let $\MDP$ a MDP and consider the $\lambda$-penalized MDP $\MDP(\lambda)$. Then $\MDP$ is indexable if and only if there exist $S$ thresholds $\mu_1, \dots, \mu_S \in \mathbb{R}$ and $S+1$ policies $\pi_0, \dots, \pi_S$ such that:
    \begin{enumerate}
        \item \emph{(Policies motonony)} $\pi_0 = \calS, \pi_S = \emptyset$, and $\forall 0 \leq i \leq S-1, \pi_{i+1} \subset \pi_i$
        \item \emph{(Penalties monotonicity)} $\forall 1 \leq i \leq S - 1, \mu_i \leq \mu_{i+1}$,
        \item \emph{(Optimality)} With $\mu_0 := -\infty$ and $\mu_{S+1} := +\infty$, then $\forall 0 \leq i \leq S \text{ and } \forall \lambda \in ( \mu_{i}, \mu_{i+1} ), \pi_i$ is the only BO policy in $\MDP(\lambda)$.
    \end{enumerate}
\end{lemma}
For $0 \leq i \leq S-1$ the set ${\pi_{i} \setminus \pi_{i+1}}$ represent the states in which it becomes better to choose action $0$ over action $1$ when $\lambda$ grows bigger than $\mu_i$. For all $s \in \pi_{i} \setminus \pi_{i+1}$ the \emph{Whittle index of state $s$} is then $\lambda_s := \mu_i$.

A last characterization is to use the activation advantage. Recalling that $\MDP$ is unichain, one can define the optimal activation advantage function $\alpha_s^*(\lambda)$ as the activation advatage under a BO policy $\pi^*$, which is uniquely defined as:
\begin{equation}
    \label{eq:act_adv_optimal}
    \alpha^*_s(\lambda) := \alpha_s^{\pi^*}(\lambda) = r^1_s - r^0_s + \big( P^1_{s,\cdot} - P^0_{s,\cdot} \big) \cdot b^*(\lambda),
\end{equation}
where $b^*(\lambda)$ is any solution to Equation (\ref{eq:BO}) for the $\lambda$-penalized MDP $\MDP(\lambda)$. Then, assuming that $\MDP$ is indexable the Whittle index of state $s$ is the only penalty $\lambda_s \in \mathbb{R}$ that satisfies $\alpha^*_s(\lambda_s)=0$ (note: for unichain MDPs, there always exists such a value\blue{, see \cite[Corollary 3]{gast2023testing}}).

\paragraph{Characteristics of the activation advantage function}\label{par:adv_is_continuous}
On each interval $(\mu_i, \mu_{i+1})$ of Lemma \ref{lem:characterization_WI}, the only BO policy in $\MDP(\lambda)$ is $\pi_i$, so by definition of the optimal activation advantage function \blue{(\ref{eq:act_adv_optimal})}, for all $0 \leq i \leq S - 1$ and $s \in \calS, \forall \lambda \in (\mu_i, \mu_{i+1}): \alpha^*_s (\lambda) = \alpha^{\pi_i}_s (\lambda).$

As a consequence, for every state $s \in \calS$, $\alpha^*_s(\lambda)$ is piecewise affine in $\lambda$, with $S+1$ pieces. From \cite[Lemma 2.(ii)]{gast2023testing} it is also continuous\footnote{Note that some papers on learning Whittle index \cite{nakhleh2021neurwin} also introduced a notion of strict indexability that imposes the function $\lambda\mapsto\alpha^*_s: \lambda \mapsto \alpha_s^*(\lambda)$ to be nonincreasing. This notion is strictly stronger and our work does not require that.}.

\paragraph*{Illustration} Let us take the unichain MDP from Figure \ref{fig:index_ex}: it is known as the restart problem \cite{avrachenkov2022whittle}. Panel (\ref{fig:index_ex:passive}) represents transition probabilities and rewards. For the passive action, transitions are random, whereas for the active action all transitions lead deterministically to state $0$ and no reward is earned. In Panel (\ref{fig:index_ex:panel_adv}) we plot the optimal activation advantage function in terms of the penalty $\lambda$. The corresponding BO policies in the $\lambda$-penalized MDP are indicated at the top of the figure. We observe that each function crosses the x-axis exactly once, so this MDP is indexable and we can read the Whittle indices: $\lambda_0=-0.9$, $\lambda_1\approx-0.73$, $\lambda_2\approx-0.51$, $\lambda_3\approx-0.26$ and $\lambda_4\approx0.01$.

\begin{figure}[hbtp]
    \begin{subfigure}[t]{0.42\textwidth}
        \centering
        \input{IPE_figures/indexable_MDP_ex_act_and_pass.tex}
        \caption{Transitions and rewards. For action $0$: probability $0.9$ for fat dashed arrows, and $0.1$ for normal dashed ones. For action $1$: all transitions (full) are deterministic.}    
        \label{fig:index_ex:passive}
    \end{subfigure}
    % \begin{subfigure}[t]{0.275\textwidth}
    %     \centering
    %     \input{IPE_figures/indexable_MDP_ex_active.tex}
    %     \caption{Transitions and rewards of action $1$.}
    %     \label{fig:index_ex:active}
    % \end{subfigure}    
    \begin{subfigure}[t]{0.575\textwidth}
        \centering
        \includesvg[width=0.95\textwidth]{Plots/ex_indexable.svg}
        \caption{Activation advantage function of the BO policy in $\MDP(\lambda)$ in terms of $\lambda$.}
        \label{fig:index_ex:panel_adv}
    \end{subfigure}
    \caption{An indexable MDP and the computation of its Whittle indices.}
    \Description{Illustration of indexability: indexable 5-states MDP of which we compute the Whittle }
    \label{fig:index_ex}
\end{figure}

%% file: IPE_figures/indexable_MDP_ex_act_and_pass.tex
\tikzstyle{ipe stylesheet} = [
  ipe import,
  even odd rule,
  line join=round,
  line cap=butt,
  ipe pen normal/.style={line width=0.4},
  ipe pen heavier/.style={line width=0.8},
  ipe pen fat/.style={line width=1.2},
  ipe pen ultrafat/.style={line width=2},
  ipe pen normal,
  ipe mark normal/.style={ipe mark scale=3},
  ipe mark large/.style={ipe mark scale=5},
  ipe mark small/.style={ipe mark scale=2},
  ipe mark tiny/.style={ipe mark scale=1.1},
  ipe mark normal,
  /pgf/arrow keys/.cd,
  ipe arrow normal/.style={scale=7},
  ipe arrow large/.style={scale=10},
  ipe arrow small/.style={scale=5},
  ipe arrow tiny/.style={scale=3},
  ipe arrow normal,
  /tikz/.cd,
  ipe arrows, % update arrows
  <->/.tip = ipe normal,
  ipe dash normal/.style={dash pattern=},
  ipe dash dotted/.style={dash pattern=on 1bp off 3bp},
  ipe dash dashed/.style={dash pattern=on 4bp off 4bp},
  ipe dash dash dotted/.style={dash pattern=on 4bp off 2bp on 1bp off 2bp},
  ipe dash dash dot dotted/.style={dash pattern=on 4bp off 2bp on 1bp off 2bp on 1bp off 2bp},
  ipe dash normal,
  ipe node/.append style={font=\normalsize},
  ipe stretch normal/.style={ipe node stretch=1},
  ipe stretch normal,
  ipe opacity 10/.style={opacity=0.1},
  ipe opacity 30/.style={opacity=0.3},
  ipe opacity 50/.style={opacity=0.5},
  ipe opacity 75/.style={opacity=0.75},
  ipe opacity opaque/.style={opacity=1},
  ipe opacity opaque,
]
\begin{tikzpicture}[ipe stylesheet]
  \node[ipe node, anchor=base]
     at (120, 676) {$r^0 = [0.9, 0.9^2, 0.9^3, 0.9^4, 0.9^5]$};
  \draw[ipe dash dashed, ->]
    (76, 768)
     -- (108, 792);
  \draw[ipe dash dashed, ->]
    (108, 800)
     .. controls (88, 816) and (112, 832) .. (116, 808);
  \draw[ipe pen fat, ipe dash dashed, ->]
    (144, 700)
     -- (96, 700);
  \draw[ipe pen fat, ipe dash dashed, ->]
    (76, 708)
     -- (68, 748);
  \draw[ipe pen fat, ipe dash dashed, ->]
    (172, 748)
     -- (164, 708);
  \draw[ipe pen fat, ipe dash dashed, ->]
    (64, 772)
     .. controls (52, 796) and (84, 796) .. (72, 772);
  \draw[ipe dash dashed, ->]
    (164, 768)
     -- (132, 792);
  \draw[ipe dash dashed, ->]
    (84, 712)
     -- (112, 788);
  \draw[ipe dash dashed, ->]
    (156, 712)
     -- (128, 788);
  \draw[->]
    (132, 800)
     .. controls (152, 816) and (128, 832) .. (124, 808);
  \node[ipe node, anchor=base]
     at (120, 664) {$r^1 = [0, 0, 0, 0, 0]$};
  \draw[->]
    (80, 764)
     -- (112, 788);
  \draw[->]
    (84, 700)
     -- (116, 788);
  \draw[->]
    (156, 700)
     -- (124, 788);
  \draw[->]
    (160, 764)
     -- (128, 788);
  \draw[ipe pen fat, ipe dash dashed, ->]
    (132, 796)
     .. controls (152, 804) and (168, 792) .. (168, 772);
  \filldraw[fill=white]
    (156, 700) circle[radius=12];
  \node[ipe node, anchor=center]
     at (156, 700) {$\mathbf{2}$};
  \filldraw[fill=white]
    (84, 700) circle[radius=12];
  \node[ipe node, anchor=center]
     at (84, 700) {$\mathbf{3}$};
  \filldraw[fill=white]
    (68, 760) circle[radius=12];
  \node[ipe node, anchor=center]
     at (68, 760) {$\mathbf{4}$};
  \filldraw[fill=white]
    (172, 760) circle[radius=12];
  \node[ipe node, anchor=center]
     at (172, 760) {$\mathbf{1}$};
  \filldraw[fill=white]
    (120, 796) circle[radius=12];
  \node[ipe node, anchor=center]
     at (120, 796) {$\mathbf{0}$};
\end{tikzpicture}

%% file: sections/topology_indexable_arms.tex
In this section we investigate the tricky topology of unichain, indexable MDPs. Given some transition matrices $(P^a)_{a \in \calA}$ and reward vectors $(r^a)_{a \in \calA}$, there exist efficient algorithms to test indexability and compute the Whittle indices of a unichain MDP. The fastest known is \WISC~\cite{gast2023testing}, an efficient implementation of which can be found \href{https://gitlab.inria.fr/markovianbandit/markovianbandit}{here}\footnote{https://gitlab.inria.fr/markovianbandit/markovianbandit}. Hence, a natural algorithm to learn Whittle indices would be to use past observations to construct estimates $\hat{P}$ of the transition matrices and to return (as an estimate of the true Whittle indices) the Whittle indices of the estimated MDP $\hatMDP := (\calS, \calA, \hat{P},r)$. In Theorem \ref{th:non-indexable_neighborhood} below we show that such a ``naive'' approach does not work in general because indexability is not robust to estimation noise: $\hatMDP$ is not necessarily indexable no matter how close it is to the true MDP $\MDP$. This is why in the next Section \ref{sec:EWISC} we introduce \EWISC, a new algorithm which refines \WISC~and outputs a vector of indices \EWISC$(\hatMDP)$ that converges to the true indices \WISC$(\MDP)$ as $\hatMDP$ converges to $\MDP$, without requiring the estimate MDP $\hatMDP$ to be indexable. In Theorem \ref{th:distinct_WI} we show that if an MDP has distinct Whittle indices then it has an indexable neighborhood.

\subsection{Preliminaries}

First, let us introduce some definitions and an assumption to facilitate the formal analysis of the subset of indexable MDPs. From now on and to give formal meaning to the notion of convergence, we define a distance function in the set of MDPs: given $\MDP:=\big(\calS, \calA, (P^a)_{a \in \calA}, (r^a)_{a \in \calA}\big)$ and $\hatMDP:=\big(\calS, \calA, (\hat{P}^a)_{a \in \calA}, (r^a)_{a \in \calA}\big)$ two MDPs with identical state/action spaces and rewards\footnote{Throughout the paper, we denote by $\norminf{P}$ the matrix infinity norm: $\norminf{P}:=\max_{s}\sum_{s'}|P_{s,s'}|$, and $\norminf{r}$ the vector infinity norm: $\norminf{r}:=\max_s |r_s|$}:
\begin{equation*}
    \norminf{\MDP - \hatMDP} := \max_{a \in \calA} \norminf{P^a - \hat{P}^a}.
\end{equation*}

The set of MDPs with state space $\calS$, action space $\calA$ and rewards $(r_s^a)_{a \in \calA, s \in \calS}$ together with this distance is a metric space. In the rest of this section, $\MDP:=\big(\calS, \calA, (P^a)_{a \in \calA}, (r^a)_{a \in \calA}\big)$ denotes the indexable MDP whose Whittle indices we want to learn. We recall that $\MDP$ is unichain. 

In addition, we assume that $\hatMDP$ is a sufficiently accurate estimate of $\MDP$ in the sense that every transition that exists in $\MDP$ also exists in $\hatMDP$, and vice versa. In other words, both MDPs share the \emph{same support}:
\begin{equation}
    P^a_{s,s'} = 0 \Longleftrightarrow \hat{P}^a_{s,s'} = 0. \label{eq:same_support}
\end{equation}
\blue{As we will discuss in Subsection \ref{subsec:BLINQ_learning_model}, our learning model $\BLINQ$ ensures that Equation (\ref{eq:same_support}) is satisfied as soon as every transition of $\MDP$ has been explored.}

Lastly, let us remark that the definition of a unichain MDP is purely based on the support of the transition matrices. As $\MDP$ is unichain and both $\MDP$ and $\hatMDP$ share the same support, we deduce that $\hatMDP$ is unichain as well.

\subsection{Boundary points of indexable MDPs set}

\begin{theorem}
    \label{th:non-indexable_neighborhood}
    There exists a MDP $\MDP=\big(\calS, \calA, (P^a)_{a \in \calA}, (r^a)_{a \in \calA}\big)$ that is unichain and indexable such that for all $\varepsilon>0$, there exists a unichain and non-indexable MDP $\hatMDP=\big(\calS, \calA, (\hat{P}^a)_{a \in \calA}, (r^a)_{a \in \calA}\big)$ such that $\norminf{\MDP - \hatMDP}\le\varepsilon$. 
\end{theorem}

\begin{proof}

    \blue{For every $\varepsilon \geq 0$ let us denote by $\MDP_\varepsilon$ the MDP of Figure \ref{fig:MDP_NI_cv_I}. It is unichain and by construction, the MDP sequence $\big( \MDP_\varepsilon \big)_{\varepsilon \geq 0}$ converges to $\MDP_0$ as $\varepsilon$ goes to $0$. We claim that $\MDP_\varepsilon$ is indexable iff $\varepsilon = 0$.

    \begin{figure}[ht]
        \input{IPE_figures/non_indexable_MDP_that_converges_to_indexable.tex}
        \centering
        \caption{Despite the limit MDP for $\varepsilon = 0$ being indexable, when $\varepsilon > 0$ this unichain MDP is non-indexable. Solid and dashed arrows represent the transitions corresponding to the active and passive action, respectively. Labels indicate transition probability.}
        \Description{Example of a sequence of non-indexable MDPs that has a indexable limit.}
        \label{fig:MDP_NI_cv_I}
    \end{figure} 

    We first notice that under any policy $\pi \subseteq \calS$, the set of visited states only depends on whether $\pi$ chooses the active action in state $3$ or not: if $3 \in \pi$ the induced Markov chain visits state $1$ but never visits $2$ (we call this \emph{path 1}), and if $3 \notin \pi$ it visits state $2$ but never visits $1$ (we call this \emph{path 2}).  We define by $r^\text{path 1}_{3 \rightarrow 0} (\lambda)$ the expected reward of an optimal policy that goes from state $3$ to $0$ via path 1, and $r^\text{path 2}_{3 \rightarrow 0} (\lambda)$ when going via path 2 (note that $r^\text{path 1}_{3 \rightarrow 0} (\lambda)$ and $r^\text{path 2}_{3 \rightarrow 0} (\lambda)$ do not include the reward obtained to go from state $0$ to state $3$ as it does not depend on the path). Let us compute their values. 

    \textbf{Path 1}. If action $0$ is chosen in state $3$, the decision maker gets an instant reward $1$ and transitions to state $1$. Then the decision maker stays on average $2$ time steps in state $1$ (no matter the chosen action), and obtains either $1-\lambda$ by choosing action $1$, or $1$ by choosing action $0$. This shows that:
    \begin{equation}
        \label{eq:path1}
        r^\text{path 1}_{3 \rightarrow 0} (\lambda) = 1 + 2\max(1-\lambda, 1) =
            \begin{cases}
                3 - 2 \lambda, & \text{if }\lambda \leq 0\\
                3, & \text{otherwise}.
            \end{cases}
    \end{equation}
    \textbf{Path 2}. If action $1$ is chosen is state $3$, the decision maker gets an instant reward $1-\lambda$ and transition to state $2$. In state $2$, there are two choices: choosing action $0$ induces a reward of $1$ during an average of $1/(1/2+2\varepsilon)$ time-steps, and choosing action $1$ induces a reward of $1-\lambda$ during an average of $1/(1/2-\varepsilon)$ time-steps. This shows that:
    \begin{equation}
        \label{eq:path2}
        r^\text{path 2}_{3 \rightarrow 0} (\lambda) = 1 - \lambda - \max\left( \frac{1-\lambda}{1/2+2\varepsilon}, \frac{1}{1/2-\varepsilon}\right) = 
            \begin{cases}
                %1 - \lambda + \frac{1}{1/2 + 2\varepsilon} (1 - \lambda) = 
                \big( 3 - \frac{8 \varepsilon}{1 + 4 \varepsilon} \big) (1 - \lambda), & \text{if }\lambda \leq \frac{- 6 \varepsilon}{1 - 2 \varepsilon}\\
                %1 - \lambda + \frac{1}{1/2 - \varepsilon} =: 
                3 + \frac{4 \varepsilon}{1 - 2 \varepsilon} - \lambda, & \text{otherwise},
            \end{cases}
    \end{equation}

    For any penalty $\lambda \in \mathbb{R}$, a policy is optimal iff it explores the path with the highest reward. In other words, if $r^\text{path 2}_{3 \rightarrow 0} (\lambda) > r^\text{path 1}_{3 \rightarrow 0} (\lambda)$ then an optimal policy chooses active action in state $3$, and vice versa. For $\varepsilon=0$, both paths are symmetric, and action $1$ is optimal in all states iff $\lambda \geq 0$. Therefore, $\MDP_0$ is indexable, with all its indices being equal to $0$. For $\varepsilon>0$, the transitions starting from state $2$ induce an unbalance between the two paths, and the situation is more complicated.  We illustrate this in Figure~\ref{fig:NIcvtoI_paths_comparison}(a) by plotting the rewards on both paths computed in \eqref{eq:path1} and \eqref{eq:path2} when $\varepsilon>0$. We observe that as $\lambda$ grows, it becomes successively better to choose a policy that explores path 2, then 1, then 2 again, then 1 again. This shows that  $\MDP_\varepsilon$ is not indexable for $\varepsilon>0$. In Figure~\ref{fig:NIcvtoI_paths_comparison}(b) and (c), we also plot the active advantage of all states as a function on $\lambda$ for the two cases $\varepsilon>0$ (panel (b)) and $\varepsilon=0$ (panel (c)). We observe that when $\varepsilon>0$, the active advantage of state $3$ crosses the $x$-axis  multiple times which indicates that the MDP is not indexable. 

    % If the decision maker chooses action $1$ in state $2$, it gets an instant reward of $1-\lambda$ and transitions to $0$ with higher probability than if it had chosen action $0$ (which provides an instant reward of $1$). For all penalties, state $0$ provides less instant reward than state $2$ and as a consequence:
    % \begin{itemize}
    %     \item when $\lambda \geq 0$, choosing action $0$ over $1$ in state $2$ allows both a higher instant reward to be earned and a slower transition to $0$ to be made,
    %     \item when $\lambda < 0$, the instant reward associated with action $1$ is slightly higher, but the transition to the less rewarding state $0$ is slightly quicker so action $1$ is better when $|\lambda|$ is not too small. In fact: action $1$ is optimal in state $2$ iff $\lambda \leq \frac{- 6 \varepsilon}{1 - 2 \varepsilon}$,
    % \end{itemize}
    % and similarly as before one can deduce:
    % \begin{equation*}
    %     r^\text{path 2}_{3 \rightarrow 0} (\lambda) = 
    %         \begin{cases}
    %             1 - \lambda + \frac{1}{1/2 + 2\varepsilon} (1 - \lambda) = \big( 3 - \frac{8 \varepsilon}{1 + 4 \varepsilon} \big) (1 - \lambda), & \text{if }\lambda \leq \frac{- 6 \varepsilon}{1 - 2 \varepsilon}\\
    %             1 - \lambda + \frac{1}{1/2 - \varepsilon} =: 3 + \frac{4 \varepsilon}{1 - 2 \varepsilon} - \lambda, & \text{otherwise},
    %         \end{cases}
    % \end{equation*}    

    \begin{figure}[ht]
        \begin{tabular}{ccc}
            \includesvg[width=0.4\textwidth]{Plots/NIcvtoI_paths_comparison.svg}
            & \includesvg[width=0.25\textwidth]{Plots/NI_cv_to_I_eps0_01.svg}
            & \includesvg[width=0.25\textwidth]{Plots/NI_cv_to_I_eps0.svg}\\        
            (a) Expected maximum & (b) Optimal activation & (c) Optimal activation \\ cumulated rewards ($\varepsilon>0$) & advantage ($\varepsilon=0.01$) & advantage ($\varepsilon=0$)
        \end{tabular}
        \centering
        \caption{Panel (a): expected maximum cumulated rewards earned by starting from state $3$ and finishing in state $0$. Panels (b) and (c): Optimal activation advantages computed by \EWISC~on $\MDP_\varepsilon$, for $\varepsilon = 0.01$ and $\varepsilon = 0$, respectively.}
        \label{fig:NIcvtoI_paths_comparison}
        \Description{Computation of the expected cumulated rewards associated with two paths. They have the property of crossing each other several times.}
    \end{figure}
}

    % Another interesting point of view is to run \EWISC. In Figure~\ref{fig:EWISC_on_cvMDP}, we plot the optimal activation advantage functions $\alpha^*_s(\lambda)$ as a function of $\lambda$ for three values of $\varepsilon\in\{0.01, 0.006,0\}$. When $\varepsilon=0$, the advantage function is simply $\alpha^*_s(\lambda)=-\lambda$ for all states: the MDP is indexable with all indices being equal to $0$. Yet, when $\frac{1}{2} > \varepsilon>0$, the Bellman optimal policy is $\pi := \{0, 1, 2\}$ for $\lambda \in \big( \frac{-8 \varepsilon}{1 + 4 \varepsilon}, \frac{- 6 \varepsilon}{1 + 4 \varepsilon} \big)$ and $\pi' := \{ 3 \}$ for $\lambda \in \big( 0, \frac{4 \varepsilon}{1 - 2 \varepsilon} \big)$. As $\pi' \nsubseteq \pi$ this proves $\MDP_\varepsilon$ is not indexable when $\varepsilon > 0$.
    % \begin{figure}[ht]
    %     \includesvg[width=0.325\textwidth]{Plots/NI_cv_to_I_1.svg}
    %     \includesvg[width=0.325\textwidth]{Plots/NI_cv_to_I_2.svg}
    %     \includesvg[width=0.325\textwidth]{Plots/NI_cv_to_I_3.svg}
    %     \centering
    %     \caption{In $\MDP_\varepsilon$, activation advantage of each state under BO policies as $\lambda$ varies, for $\varepsilon = 0.01, 0.006$ and $0$, respectively.}
    %     \Description{Computation of EWISC on some non-indexable MDPs antd their indexable limit.}
    %     \label{fig:EWISC_on_cvMDP}
    % \end{figure}
\end{proof}

% We just showed that the set of indexable MDPs is not open in the set of MDPs. It is however worth noting that if the indexable MDP $\MDP$ has distinct indices (that is, no pair of states has the same Whittle index) then it has an indexable neighborhood. This shall be properly formalized and proved in Theorem \ref{th:distinct_WI}.

\subsection{The case of distinct Whittle indices}

The previous Section \ref{sec:topology_I} highlighted the necessity of generalizing the \WISC~algorithm with the introduction of \EWISC~to cope with the fact that our learning-based estimated MDP has no guarantee to be indexable, no matter how accurate it is. The next theorem shows that when the indices of the original model $\MDP$ are all distinct, then there exists a neighborhood around $\MDP$ such that all models in this neighborhood are indexable. 

\begin{theorem}
    \label{th:distinct_WI}
    Let $\MDP=\big(\calS, \calA, (P^a)_{a \in \calA}, (r^a)_{a \in \calA}\big)$ be a unichain and indexable MDP such that all its Whittle indices are different from one another. Then there exits a threshold $\delta > 0$ such that any estimated MDP $\hatMDP=\big(\calS, \calA, (\hat{P}^a)_{a \in \calA}, (r^a)_{a \in \calA}\big)$ that has the same support as $\MDP$, as defined in Equation (\ref{eq:same_support}), satisfies:
    \begin{equation*}
        \norminf{\hatMDP - \MDP}\le\delta \implies \hatMDP \text{ is indexable.}
    \end{equation*}
\end{theorem}

\blue{This fact was unknown to the best of our knowledge and provides more details about the topology of indexable and unichain MDPs. It implies that when learning the Whittle indices of some unknown MDP $\MDP$ with distinct indices, if the learning agent has a sufficiently accurate approximation of $\MDP$ then it can call \WISC~instead of \EWISC. However it is hard in practice to certify that: 1. the unknown MDP $\MDP$ has distinct Whittle indices and 2. its estimate $\hatMDP$ is sufficiently accurate. Hence in the general case \EWISC~is needed regardless of whether $\MDP$ has distinct Whittle indices or not.}

The proof of Theorem \ref{th:distinct_WI} relies on the geometric interpretation of Whittle indices introduced in \textsection\ref{par:charact_of_WI} and an important result introduced in the analysis of \EWISC~in Subsection \ref{subsec:cv_of_EWISC}. Its detailed version is subsequently postponed to Appendix \ref{app:sec:proof_distinct_WI}.

\paragraph*{Sketch of the proof} We first choose a $\delta > 0$ such that for any state $s \in \calS$, the optimal activation advantage function of state $s$ in $\hatMDP$---named $\hat{\alpha}^*_s$ and defined in Equation (\ref{eq:act_adv_optimal})---cancels out only locally around $\lambda_s$: the Whittle index of state $s$ in $\MDP$. We then characterize the variations of $\hat{\alpha}^*_s$ locally around $\lambda_s$: the inflexion points of $\hat{\alpha}^*_s$ happen at penalties for which some other state optimal activation advantage function $\hat{\alpha}^*_{s'}$ cancels out. As all Whittle indices are different and each $\hat{\alpha}^*_s$ cancels out only around its associated Whittle index $\lambda_s$, it follows that $\hat{\alpha}^*_s$ has exactly one inflexion point near $\lambda_s$. Combining the former and the latter result allows us to deduce that $\hat{\alpha}^*_s$ crosses the x-axis only once on $\mathbb{R}$, thus making $\hatMDP$ indexable.

%% file: IPE_figures/non_indexable_MDP_that_converges_to_indexable.tex
\begin{tikzpicture}[ipe import]
  \draw[->]
    (47.7943, 752.871)
     .. controls (80, 784) and (128, 792) .. (156, 792);
  \draw[dashed, ->]
    (44, 748)
     .. controls (80, 744) and (128, 764) .. (156, 784);
  \draw[dashed, ->]
    (40, 756)
     .. controls (44, 784) and (8, 772) .. (28, 748);
  \node[ipe node, anchor=center]
     at (83.909, 785.121) {$\frac{1}{2} $};
  \node[ipe node, anchor=center]
     at (111.909, 753.121) {$\frac{1}{2} $};
  \node[ipe node, anchor=center]
     at (19.909, 773.121) {$\frac{1}{2} $};
  \draw[->]
    (40, 732)
     .. controls (44, 704) and (8, 716) .. (28, 740);
  \node[ipe node, anchor=center]
     at (19.909, 717.121) {$\frac{1}{2} $};
  \draw[dashed, ->]
    (156, 696)
     -- (52, 740);
  \draw[->]
    (180, 696)
     -- (284, 740);
  \node[ipe node, anchor=center]
     at (112, 720) {$1$};
  \node[ipe node, anchor=center]
     at (224, 720) {$1$};
  \node[ipe node, anchor=center]
     at (219.909, 753.121) {$\frac{1}{2} - \varepsilon$};
  \node[ipe node, anchor=center]
     at (259.909, 785.121) {$\frac{1}{2} +2 \varepsilon$};
  \draw[->]
    (296, 732)
     .. controls (292, 704) and (328, 716) .. (308, 740);
  \node[ipe node, anchor=center]
     at (327.909, 717.121) {$\frac{1}{2} - 2 \varepsilon$};
  \draw[dashed, ->]
    (296, 756)
     .. controls (292, 784) and (328, 772) .. (308, 748);
  \node[ipe node, anchor=center]
     at (323.909, 773.121) {$\frac{1}{2} +
 \varepsilon$};
  \filldraw[fill=white]
    (168, 788) circle[radius=12];
  \filldraw[fill=white]
    (168, 696) circle[radius=12];
  \node[ipe node, anchor=center]
     at (167.794, 787.35) {$\mathbf{0}$};
  \node[ipe node, anchor=center]
     at (167.794, 695.35) {$\mathbf{3}$};
  \draw[dashed, ->]
    (44, 748)
     .. controls (80, 744) and (128, 764) .. (156, 784);
  \draw[dashed, ->]
    (292, 748)
     .. controls (256, 744) and (208, 764) .. (180, 784);
  \draw[->]
    (288, 752)
     .. controls (256, 784) and (208, 792) .. (180, 792);
  \filldraw[fill=white]
    (296, 744) circle[radius=12];
  \node[ipe node, anchor=center]
     at (295.795, 743.35) {$\mathbf{2}$};
  \node[ipe node, anchor=center]
     at (168, 676) {$r^0 = r^1 = [0, 1, 1, 1]$};
  \draw[->]
    (172, 776)
     .. controls (188, 756) and (188, 732) .. (172, 708);
  \draw[dashed, ->]
    (164, 776)
     .. controls (148, 756) and (148, 732) .. (164, 708);
  \node[ipe node, anchor=center]
     at (148, 744) {$\frac{1}{2}$};
  \node[ipe node, anchor=center]
     at (188, 744) {$\frac{1}{2}$};
  \draw[->]
    (172, 800)
     .. controls (168, 828) and (200, 816) .. (180, 792);
  \draw[dashed, ->]
    (164, 800)
     .. controls (168, 828) and (136, 816) .. (156, 792);
  \node[ipe node, anchor=center]
     at (156, 824) {$\frac{1}{2}$};
  \node[ipe node, anchor=center]
     at (180, 824) {$\frac{1}{2}$};
  \filldraw[fill=white]
    (40, 744) circle[radius=12];
  \node[ipe node, anchor=center]
     at (39.794, 743.35) {$\mathbf{1}$};
\end{tikzpicture}

%% file: sections/EWISC.tex
Theorem \ref{th:non-indexable_neighborhood} showed that an arbitrarily close neighbor of an indexable MDP is not necessarily indexable. For our model-based approach, this poses a problem: the estimated MDP learned by the decision maker might always be non-indexable, even if the true MDP is indexable and the learning time is arbitrarily large. Hence we introduce in this section a new algorithm, that we call \EWISC for \texttt{E}xtended \WISC. It returns a set of indices even for non-indexable inputs. As we will see later, if a sequence of MDPs $\hatMDP$ converges to an indexable MDP $\MDP$, then the values returned by $\EWISC(\hatMDP)$~converges to the true indices of $\MDP$. 

\EWISC~works similarly as \WISC\cite{gast2023testing}: it computes the optimal activation advantages from Equation \ref{eq:act_adv_optimal} as functions of the penalty. The main difference between \EWISC~and \WISC~is that for each state $s \in \calS$, the former returns the \emph{average} of all penalties $\lambda$ s.t. $\alpha^*_s(\lambda) = 0$, whereas the latter returns the \emph{only} penalty $\lambda$ s.t. $\alpha^*_s(\lambda) = 0$, which makes sense only if $\MDP$ is indexable.

\subsection{High level description}

% \paragraph{High level description}
Let $\MDP$ be a unichain MDP. By \cite[Lemma II.(iii)]{gast2023testing} the BO policy in the $\lambda$-penalized MDP $\MDP(\lambda)$ is unique except when $\lambda$ coincides with a root of some optimal activation advantage function $\alpha_s^*$.

\blue{When $\MDP$ is indexable, the BO policy $\pi$ in $\MDP(\lambda)$ is decreasing in $\lambda$, as characterized in Lemma \ref{lem:characterization_WI}. In that case, \WISC~works by deleting states one by one from the initial policy $\pi = \calS$, which is BO for $\lambda=-\infty$. It does so until it reaches $\pi = \emptyset$, which is BO for $\lambda=-\infty$. Because of the monotonicity, this is achieved after $S$ deletions. When $\MDP$ is not indexable, the BO policy as a function of the penalty $\lambda$ is not monotonic anymore. \EWISC~accounts for that by introducing the possibility to add one state back to $\pi$ at each step. The end condition of \EWISC~is $\pi = \emptyset$, just as \WISC.}

More formally, \EWISC~computes a sequence of penalties $\mu_1, \dots, \mu_K$ and policies $\pi_0, \dots, \pi_K$ s.t.
\label{par:charact_of_EWISC}
\begin{enumerate}
    \item \emph{(Policies ``motononicity'')} $\pi_0 = \calS, \pi_K = \emptyset$,
    \item \emph{(Penalties monotonicity)} $\forall 1 \leq i \leq K - 1, \mu_i \leq \mu_{i+1}$,
    \item \emph{(Optimality)} Defining $\mu_0 := -\infty$ and $\mu_{K+1} := +\infty$, then $\forall 0 \leq i \leq S \text{ and } \forall \lambda \in ( \mu_{i}, \mu_{i+1} ), \pi_i$ is the only BO policy in $\MDP(\lambda)$.
\end{enumerate}
\blue{where $K$ denotes the number of additions or deletions performed by \EWISC. Note that $K \geq \calS$, with equality iff $\MDP$ is indexable. Also, since \EWISC~never explores two identical policies, $K \leq | \calA^{\calS} | = 2^S$. The question of whether or not this bound is tight remains open to our knowledge.} The difference between the above characterization and the characterization of Whittle indices of Lemma \ref{lem:characterization_WI} in \textsection \ref{par:charact_of_WI} is that $S$ is replaced by $K \geq S$, and consecutive policies do not need to evolve in a monotonic fashion anymore. \EWISC~then returns a $S$-sized vector where each coordinate is the average of all roots of $\alpha_s^*$. A detailed pseudocode of \EWISC~is given by Algorithm \ref{alg:EWISC} in Appendix \ref{app:sec:EWISC}. It is inspired and refined from \WISC's pseudocode \cite[Algorithm 1]{gast2023testing}. As proved in the same appendix, \EWISC's time complexity is $O(K S^2)$.

\subsection{Linear convergence of \EWISC}
\label{subsec:cv_of_EWISC}

Let $\hatMDP:=\big(\calS, \calA, (\hat{P}^a)_{a \in \calA}, (r^a)_{a \in \calA}\big)$ denote a MDP whose rewards are identical to the ones of $\MDP$. In addition, we suppose that $\hatMDP$ is an accurate estimate of $\MDP$:
\begin{assumption}
    \label{ass:tildeM_close}
    $\hatMDP$ is sufficiently close to $\MDP$, in the sense that:
    \begin{equation*}
        \norminf{\MDP-\hatMDP} \leq 1 / \max_{\pi \in \calA^{\calS}} \diam{P^\pi} \text{ and } \norminf{\MDP-\hatMDP} \leq \frac{1}{2} \min_{\lambda_s \neq \lambda_{s'}} |\lambda_s - \lambda_{s'}|,
    \end{equation*}
    where $(\lambda_s)_{s \in \calS}$ are the Whittle indices of $\MDP$ and $\diam{P^\pi}$ is the diameter\footnote{The diameter of a unichain transition matrix $P^\pi$ is the maximum time needed to go from any state to a state in its unique recurrent class. Denoting by $\rand{\tau_{s,s'}}$ the random variable representing the number of time steps necessary to go from state $s$ to state $s'$ we define $\diam{P^\pi} := \max_{s \in \calS,s' \in \calS_r} \mathbb{E} [\rand{\tau_{s,s'}}]$, where $\calS_r$ is the (unique) recurrent class of $P^\pi$. } of the transition matrix $P^\pi$.
\end{assumption}

The following theorem formalizes a linear convergence of the output of \EWISC~on input $\hatMDP$ to the real Whittle indices of $\MDP$. We recall the following notation:
\begin{itemize}
    \item $\norminf{\MDP - \hatMDP} := \max_{a \in \calA} \norminf{P^a - \hat{P}^a}$ is the distance between $\MDP$ and $\hatMDP$,
    \item $(\lambda_s)_{s \in \calS}$ are the Whittle indices of $\MDP$.
\end{itemize}

\begin{theorem}
    \label{th:linear_conv}
    Let $\MDP$ be an indexable and unichain MDP. There exists a constant $c_{\MDP}$ that only depends on $\MDP$ such that if $\hatMDP$ is a sufficiently accurate estimate of $\MDP$ as supposed in Assumption \ref{ass:tildeM_close} and both MDPs share the same support---see Equation (\ref{eq:same_support})---then
    \begin{equation*}
        \forall s \in \calS, \big| \mathEWISC(\hatMDP)_s - \lambda_s \big| \leq c_{\MDP} \norminf{\MDP - \hatMDP}.
    \end{equation*}
\end{theorem}

\blue{\begin{remark}
    In fact, the proof of Theorem \ref{th:linear_conv} works when \EWISC$(\hatMDP)_s$ is replaced by any arbitrary value between the smallest and biggest root of $\alpha^*_s$. For instance in the last line of Algorithm \ref{alg:EWISC}, one could replace \texttt{average} with \texttt{median}, \texttt{min} or \texttt{max}. Using these values, we could also define several distinct index policies for non-indexable MDPs, but the question of what choice would be the most relevant in that case is open to our knowledge.
\end{remark}}

The proof of this theorem relies on a geometric interpretation of Whittle indices as discussed in \textsection \ref{par:charact_of_WI}. It makes use of several lemmas and corollaries in order to bound the difference between the optimal activation advantages functions computed by $\EWISC(\hatMDP)$ on one hand, and $\WISC(\MDP)$ on the other hand, uniformly across the full range of Whittle indices of $\MDP$. 

First, let us reformulate it so that it can more easily be used in the proof of Theorem \ref{th:distinct_WI}. By definition, \EWISC~returns for every state $s \in \calS$ the average of all penalties $\hat{\lambda} \in \mathbb{R}$ such that $\hat{\alpha}_s^*(\hat{\lambda}) = 0$. Therefore Theorem \ref{th:linear_conv} is a direct corollary of \ref{lem:proof_th_lin_conv:zero} in the following lemma:
\begin{lemma}
    \label{lem:proof_th_lin_conv}
    Let $\MDP$ be an indexable and unichain MDP. Then there exists constants $0 \leq \delta_{\MDP} \leq \frac{1}{2} \min_{\lambda_s \neq \lambda_{s'}} | \lambda_s - \lambda_{s'} |$ and $0 \leq c_{\MDP}$ that only depends on $\MDP$ such that if $\hatMDP$ is a sufficiently accurate estimate of $\MDP$ as supposed in Assumption \ref{ass:tildeM_close} and both MDPs share the same support---see Equation (\ref{eq:same_support})---, then for every state $s \in \calS$:
    \begin{enumerate}[label=(\roman*)]
        \item \label{lem:proof_th_lin_conv:uniqueBO} suppose that $\norminf{\hatMDP - \MDP} \leq \delta_{\MDP}$, and let $(s,s')$ be a pair of states with consecutive and distinct Whittle indices, i.e. $\lambda_s < \lambda_{s'}$ and no other state $s'' \in \calS$ satisfy $\lambda_s < \lambda_{s''} < \lambda_s'$. Then for any $\lambda \in [ \lambda_s + \delta_{\MDP}, \lambda_{s+1} - \delta_{\MDP} ]$ the BO policy in the $\lambda$-penalized MDP $\hatMDP (\lambda)$ is unique and is the same as in $\MDP(\lambda)$,
        \item \label{lem:proof_th_lin_conv:zero} there exists at least one penalty $\hat{\lambda} \in \mathbb{R}$ such that $\hat{\alpha}^*_s(\hat{\lambda}) = 0$, and all such penalties satisfy $| \lambda_s - \hat{\lambda} | \leq c_{\MDP} \norminf{\MDP - \hatMDP}.$ 
    \end{enumerate} 
\end{lemma}

The detailed proof of Lemma \ref{lem:proof_th_lin_conv} is provided in Appendix \ref{app:sec:proof_lem2}.

\paragraph{Sketch of proof of Lemma \ref{lem:proof_th_lin_conv}} 
\label{par:sketch_proof_lem2} The main difficulty of the proof lies in the process of finding an upper bound to the \emph{optimal activation advantage gap}---that is, the difference between optimal activation advantages functions in $\hatMDP$ and $\MDP$---, for every ``relevant'' penalty $\lambda$. In other words, we need to bound
\begin{equation*}
    \sup_{\text{relevant penalty } \lambda} \norminf{\hat{\alpha}^* (\lambda) - \alpha^*(\lambda)}
\end{equation*}
where $\norminf{\cdot}$ designates the classical infinite-norm of a vector, i.e. $\norminf{\alpha} := \max_s |\alpha_s|$.

At first, Lemma~\ref{lem:WI_bounded} gives meaning to what we consider as relevant penalties and shows that one can restrict to a closed interval of penalties. Next, Lemma \ref{lem:conv_of_adv} bounds the difference between the activation advantage of any state under any policy in $\hatMDP$ on one hand, and $\MDP$ on the other hand. We emphasize that the considered state and policy should be the same in both MDPs. Then, Lemma \ref{lem:uniqueBO_close} states that if the MDPs are sufficiently close then the BO policy is the same in both. These two lemmas allow us to bound the difference between the optimal activation advantages functions in $\hatMDP$ and $\MDP$ for penalties that are not too close to the Whittle indices of $\MDP$. After that, Lemma \ref{lem:bound_slope_adv} provide us with a way to bound the optimal activation advantage gap for penalties close to the Whittle indices of $\MDP$. Finally, Lemma \ref{lem:technical_cv_fct} applied to the optimal advantages functions of both MDPs will conclude the proof. 

\subsection{Extension to discounted MDPs}
\label{subsec:discountedEWISC}

Similarly to \WISC\cite[Section 7]{gast2023testing}, \EWISC~can be easily extended to the case of discounted MDPs. Given a \emph{discount factor} $\beta \in (0,1)$, we consider the problem of finding an optimal policy as introduced in \textsection\ref{subsec:BO_and_advantage} with a different optimality criterion: for a policy $\pi$ and an initial state $s$, we replace the infinite-horizon average gain $g^\pi_s$ by the total discount reward $v_s^\pi$, called \emph{value}:
\begin{equation*}
    \forall s \in \calS, v_s^\pi := \mathbb{E} \bigg[ \sum_{t=0}^{\infty} \beta^t r^{\indicPi{S_t}}_{S_t} \bigg],
\end{equation*}
where $(S_t)_t$ is a random walk in the Markov chain induced by $\pi$ in $\MDP$ with $S_0 := s$ a.s. The vector $v^\pi$ satisfies Bellman's equation, that is:
\begin{equation*}
    \forall s \in \calS, v_s^\pi = r_s^{\pi_i} + \beta \sum_{s' \in \calS} P_{s,s'} v^\pi_{s'}.
\end{equation*}

Then $\pi$ is said to be \emph{optimal}\footnote{The different notions of gain optimality, Bellman optimality and $n$-bias optimality found in the literature for non-discounted MDPs collapse into a single notion of optimality for discounted MDPs\cite{gast2023testing}.} if $v_s^\pi \geq v_s^{\pi'}$ for all policies $\pi'$. The Whittle index policy for discounted MDPs is defined and characterized in the exact same way as in \textsection\ref{par:charact_of_WI} with Bellman optimality replaced by the notion of optimality for discounted MDPs, and the activation advantage of state $s$ under policy $\pi$ defined as
\begin{equation}
    \alpha^\pi_s := r^1_s - r^0_s + \beta (P^1_{s,\cdot} - P^0_{s,\cdot}) \cdot v^\pi. \label{eq:activation_adv_discounted}
\end{equation}

Just as for the non-discounted case, $\pi$ is optimal iff $\alpha^\pi_s \geq 0$ for all states $s \in \pi$ and $\alpha^\pi_s \leq 0$ for all states such that $s \notin \pi$, therefore \EWISC~works with the exact same pseudocode as in Algorithm \ref{alg:EWISC} but with the activation advantage defined in Equation (\ref{eq:activation_adv_discounted}). To compute it in the $\lambda$-penalized MDP $\MDP(\lambda)$, we simply replace Equation (\ref{eq:advantage_computation}) by (\ref{eq:activation_adv_discounted}) with $v^\pi = (I - \beta P^{\pi})^{-1} (r^{\pi} - \lambda \texttt{bin}^{\pi} )$. Note that $I - \beta P^{\pi}$ is invertible because $P^{\pi}$ is stochastic and $0 < \beta < 1$. The time complexity of \EWISC~is therefore the same for $\beta-$discounted and non-discounted MDPs.

%% file: sections/BLINQ.tex
When learning an estimated MDP $\hatMDP$ of an unknown MDP $\MDP$, Section \ref{sec:EWISC} allowed us to introduce an algorithm called \EWISC~such that $\mathEWISC(\hatMDP)$ converges to the Whittle indices of $\MDP$ regardless of the indexability of $\hatMDP$. In this section we introduce a learning framework called \BLINQ, for \texttt{B}LINQ \texttt{L}earning \texttt{I}s \texttt{N}ot \texttt{Q}-learning. It uses \EWISC~ to learn the Whittle indices of $\MDP$. We analyze it by providing performance guarantees on it, i.e. sufficient conditions on the learning time to have an accurate approximation.

\subsection{Learning model}
\label{subsec:BLINQ_learning_model}

\blue{Our learning environment is a classical reinforcement learning setup. We have access to a simulator of the MDP, that is: at time $t$, the learning agent observes the state of the MDP $\rand{S_t}$ and chooses an action $\rand{A_t} \in \calA$. The MDP then generates a reward and transitions to its next state, both being observed by the agent. This is called a navigating model, in contrast to a generative model where the agent decides which state $\rand{S_t}$ it explores at time $t$; and subsequently decides the action $\rand{A_t}$ and observes the generated reward as well as the next state. Thus the generative model is less constraining than the navigating model. We chose the latter, as it is usually the case in the literature regarding index learning \cite{robledo_borkar2022deeplearning,avrachenkov2022whittle,dhankar2025tabulardeepreinforcementlearning,jonah2026WIQL}.}

For a state $s\in\calS$ and an action $a\in\calA$, we denote by $\rand{N_t(s,a)}$ the number of time-steps when action $a$ was chosen in state $s$ before time $t$, and we denote by $\rand{N_t(s,a,s')}$ the number of times this generated a transition to state $s'\in\calS$ before time $t$. It should be clear that $\sum_{s\in\calS, a\in\{0,1\}}\rand{N_t(s,a)} = t$ and $\sum_{s'\in\calS}\rand{N_t(s,a,s')} = \rand{N_t(s,a)}$.
We define the following empirical estimate for the transition matrices whenever $\rand{N_t(s,a)}>0$:
\begin{equation}
    \label{eq:approx_MDP}
    \big[ \hat{P}^a_{s,s'} \big]_t := \frac{\rand{N_t(s,a,s')}}{\rand{N_t(s,a)}},
\end{equation}
and we denote by $\hatMDP_t:=(\calS, \calA, \big[ \hat{P} \big]_t,r)$ the estimated MDP  at time $t$. This definition ensures that Eq (\ref{eq:same_support}) is satisfied as soon as all the state-action pairs have been visited. We also recall that the rewards are known and deterministic. 

\subsection{Pseudocode} 

At each time step $t \geq 1$, we select randomly an action $\rand{A_t}\in \{0,1\}$ with uniform probability (to explore uniformly matrices $P^0$ and $P^1$). Meanwhile, we construct $\hatMDP_t$ the empirical estimate of the MDP, whose formal definition is in (\ref{eq:approx_MDP}). We then run the algorithm \EWISC~on $\hatMDP_t$.

To reduce the computation cost, we progressively space out consecutive runs of \EWISC, so that most of the time steps involve only updating $\rand{N_t(S_t,A_t)}$ and $\rand{N_t(S_t,A_t,S_{t+1})}$. As in our numerical experiments in Section \ref{sec:numerical}, one simple way to do that is to choose some factor \texttt{k} $> 1$ and to multiply the number of time steps between each consecutive runs of \EWISC~by \texttt{k} whenever \EWISC~is ran. In the pseudocode given by Algorithm \ref{alg:BLINQ}, this is represented as an oracle \texttt{update\_indices(t)} that returns \texttt{True} iff \EWISC~is ran at time \texttt{t}.

In our pseudocode, the first run of \EWISC~occurs once every pair has been visited at least once, i.e. $N_t(a,s) > 0 \forall a,s$. One can do differently, for instance one could build a restricted estimate of $\MDP$ that only contains the probability estimates and rewards of visited states and actions.

\begin{algorithm}
    \caption{: \textbf{B}LINQ \textbf{L}earning \textbf{I}s \textbf{N}ot \textbf{Q}-learning }\label{alg:BLINQ}
    \begin{algorithmic}
    \At \textbf{ time} $t = 0$:
        \State $\rand{N_0(a,s)}, \rand{N_0(a,s,s')} \gets 0$ for every states $s, s' \in \calS$ and action $a \in \calA$
        \State Start at some state $\rand{S_0}$
    \End
    \At \textbf{ time} $t \geq 1$:
        \State Choose an action $\rand{A_t}$ uniformly in $\{0, 1\}$
        \State Apply action $\rand{A_t}$ and transition to a new state $\rand{S_{t+1}}$ while earning a reward $\rand{R_t}$
        % \If{$\rand{N_t(S_t,A_t)} = 0$}
        %     \State $r^\rand{A_t}_{\rand{S_t}} \gets \rand{R_t}$ % pas nécessaire car les rewards sont connus
        % \EndIf
        \State Update $\rand{N_{t+1}(S_t,A_t)}$ and $\rand{N_{t+1}(S_t,A_t,S_{t+1})}$
        \If{$\rand{N_t(s,a)} > 0$ for every $s \in \calS, a \in \calA$ and \texttt{update\_indices(t)} $=$ \texttt{True}}:
            \State Build approximate MDP $\hatMDP_t$ as in Equation (\ref{eq:approx_MDP}) and run \EWISC($\hatMDP_t$)
        \EndIf
    \End
    \end{algorithmic}
\end{algorithm}

\subsection{Learning time analysis}

Let us now focus on the learning ability of the \BLINQ~learning framework. Firstly, we need to quantify the accuracy of the estimated MDP in terms of the number of samples collected so far thanks to Lemma \ref{lem:learning_time}. Then we shall use it in combination with Theorem \ref{th:linear_conv} to quantify the accuracy of the approximated Whittle indices, thanks to Theorem \ref{th:precision_time}. 

Proving results of asymptotical convergence to Whittle indices only make sense for the states that are visited indefinitely often. Thus we assume that all states are visited indefinitely often under the exploration policy of \BLINQ:
\begin{assumption}
    \label{ass:recurrent_states}
    $\MDP$ is communicating. In other words, in the Markov chain induced by the exploration policy of $\mathBLINQ$ (whose transition kernel is $M := \frac{1}{2} \sum_{a \in \calA} P^a$), all states are recurrent.
\end{assumption}

\begin{lemma}
    \label{lem:learning_time}
    Let $\MDP$ be a unichain and communicating MDP. Denote by $\hatMDP_t$ the estimate of $\MDP$ at time $t$ built by $\mathBLINQ$ as defined in Equation (\ref{eq:approx_MDP}). Then, there exists constants $k_\MDP, l_\MDP$ that only depend on $\MDP$ s.t. for every precision $1 \geq \varepsilon > 0$ and threshold $1 \geq \delta > 0$:
    \begin{equation*}
        \text{if the time step } t \text{ is greater than } \frac{- l_\MDP}{\varepsilon^2} \log \bigg(\frac{\delta}{k_\MDP}\bigg) \text{ then } \mathbb{P} \bigg(\norminf{\MDP - \hatMDP_t} \leq \varepsilon\bigg) \geq 1 - \delta.
    \end{equation*}
\end{lemma}

To prove this result, we use repeatedly use the following lemma.
\begin{lemma}
    \label{lem:technical_a_over_b}
    Let $a,b, \hat{a}, \hat{b} > 0$ be positive real numbers, and $\frac{b}{2} \geq \varepsilon > 0$ be a precision threshold. In addition, suppose that $|a - \hat{a}| \leq \varepsilon$ and $|b - \hat{b}| \leq \varepsilon$. Then:
    \begin{equation}
        \label{eq:lem_inequality}
        \bigg| \frac{a}{b} - \frac{\hat{a}}{\hat{b}} \bigg| \leq \frac{2(a + b)}{b^2} \varepsilon
    \end{equation}
\end{lemma}

\begin{proof}
    The left-hand side of Equation~\ref{eq:lem_inequality} multiplied by $b\hat{b}$ is equal to $|a \hat{b} - \hat{a} b |$. This shows that:
    \begin{align*}
        \left|\frac{a}{b} - \frac{\hat{a}}{\hat{b}}\right| &= \frac{1}{b\hat{b}} |a \hat{b} - \hat{a} b | = \frac{1}{b\hat{b}} | a (\hat{b} - b) - (\hat{a} - a) b |.
    \end{align*}
    The result follows because by assumption $|\hat{b} - b|\le \varepsilon$, $(\hat{a} - a)\le \varepsilon$ and $b \hat{b} \geq b \big(b - \frac{b}{2} \big) = \frac{b^2}{2}$.
\end{proof}

\begin{proof}[Proof of Lemma \ref{lem:learning_time}]
    In this proof we denote by $M$ both the Markov chain induced by the exploration policy of \BLINQ~, and its transition kernel: $M := \frac{1}{2} \sum_{a \in \calA} P^a$. Denote by $(\varphi_s)_{s \in \calS}$ the stationary distribution of the Markov chain $M$: because it is unichain and recurrent, it has a unique stationary distribution and $\varphi_s > 0$ for all states $s \in \calS$.
    
    Let $1 \geq \varepsilon > 0$ be a precision, $1 \geq \delta > 0$ be a threshold and fix a time $t \geq 1$. Let $s,s' \in \calS$ be states and $a \in \calA$ be an action. Let us also define:
    \begin{equation*}
        \varepsilon_1 := \frac{\varphi_s}{4 + 4 P^a_{s,s'}} \varepsilon
    \end{equation*}

    Applying Lemma \ref{lem:technical_a_over_b} with $a := \frac{1}{2} \varphi_s P^a_{s,s'}, b := \frac{1}{2} \varphi_s, \hat{a} := \frac{\rand{N_t(s,a,s')}}{t}, \hat{b} := \frac{\rand{N_t(s,a)}}{t}$, assuming $\varepsilon_1 \leq \frac{b}{2}$ yields the following:
    \begin{equation*}
        \bigg| \frac{\rand{N_t(s,a,s')}}{\rand{N_t(s,a)}} - P^a_{s,s'} \bigg| > \varepsilon \implies \bigg| \frac{\rand{N_t(s,a,s')}}{t} - \frac{1}{2} \varphi_s P^a_{s,s'} \bigg| > \varepsilon_1 \text{ or } \bigg| \frac{\rand{N_t(s,a)}}{t} - \frac{1}{2} \varphi_s \bigg| > \varepsilon_1
    \end{equation*}
    Note that $\varepsilon_1 \leq \frac{b}{2}$ translates to $2 + 2P^a_{s,s'} \geq \varepsilon$ which is true. This implication also holds when replacing $\varepsilon_1$ with $\frac{\varphi_s}{8} \varepsilon$ since $P^a_{s,s'} \leq 1$. Therefore:
    \begin{equation*}
    \begin{split}
        \mathbb{P} \bigg( \bigg| \frac{\rand{N_t(s,a,s')}}{\rand{N_t(s,a)}} - P^a_{s,s'} \bigg| > \varepsilon \bigg) & \leq \mathbb{P} \bigg( \big| \rand{N_t(s,a,s')} - \frac{1}{2} t \varphi_s P^a_{s,s'} \big| > t \frac{\varphi_s}{8} \varepsilon \cup \big| \rand{N_t(s,a)} - \frac{1}{2} t \varphi_s \big| > t \frac{\varphi_s}{8} \varepsilon \bigg) \\
        & \leq \mathbb{P} \bigg( \big| \rand{N_t(s,a,s')} - \frac{1}{2} t \varphi_s P^a_{s,s'} \big| > t \frac{\varphi_s}{8} \varepsilon \bigg)  + \mathbb{P} \bigg( \big| \rand{N_t(s,a)} - \frac{1}{2} t \varphi_s \big| > t \frac{\varphi_s}{8} \varepsilon \bigg).
    \end{split}
    \end{equation*}

    Let us first assume that $M$ is aperiodic. This allows us to use a concentration inequality for Markov chains \cite[Theorem 3.1]{chung2012chernoffhoeffdingboundsmarkovchains}: there exists constants $c_{s,a,s'}, c_{s,a} > 0$ that does not depend on $\varepsilon, \delta$ s.t.\:
    \begin{equation}
    \begin{split}
        \label{eq:aperiodic_bound}
        \mathbb{P} \bigg( \big| \rand{N_t(s,a,s')} - \frac{1}{2} t \varphi_s P^a_{s,s'} \big| > t \frac{\varphi_s}{8} \varepsilon \bigg) & \leq c_{s,a,s'} \exp \big( - \varepsilon^2 \varphi_s P^a_{s,s'} t / 144 T_\varphi \big), \\
        \mathbb{P} \bigg( \big| \rand{N_t(s,a)} - \frac{1}{2} t \varphi_s \big| > t \frac{\varphi_s}{8} \varepsilon \bigg) & \leq c_{s,a} \exp \big( - \varepsilon^2 \varphi_s t / 144 T_\varphi \big)
    \end{split}
    \end{equation}
    where
    \begin{equation*}
        T_\varphi := \min \{ \tau ; \max_{\psi \text{ initial distribution}} \normone{\psi M^\tau - \varphi} \leq \frac{1}{4} \}
    \end{equation*}
    is the mixing time of the Markov chain $M$. When $M$ is periodic---let us say with period $\rho$---one can look at the sequence of state-action pairs every $\rho$ time steps: at time $1$, $\rho + 1$, $2 \rho + 1$, etc. This is an aperiodic Markov chain $M_1$ which means the Equation (\ref{eq:aperiodic_bound}) holds for that Markov chain instead of $M$, with a mixing time $T_{M_1}$. Let us repeat this process for each of the $\rho$ aperiodic Markov chains one can create, i.e. by looking at the state-action pair at times $(t \rho + k)_{t \in \mathbb{N}}$ for $1 \leq k \leq \rho$. We obtain $\rho$ mixing times $(T_{M_k})_{1 \leq k \leq \rho}$ and Equation (\ref{eq:aperiodic_bound}) holds with $T_\varphi := \max_{1 \leq k \leq \rho} T_{M_k}$. Finally:
    \begin{equation*}
    \begin{split}
        \mathbb{P} \bigg( \bigg| \frac{\rand{N_t(s,a,s')}}{\rand{N_t(s,a)}} - P^a_{s,s'} \bigg| > \varepsilon \bigg) \leq \big( c_{s,a,s'} + c_{s,a} \big) \exp \big( - \varepsilon^2 \varphi_s P^a_{s,s'} t / 144 T_\varphi \big).
    \end{split}
    \end{equation*}

    Notice that if $P^a_{s,s'} = 0$, then the transition from state $s$ to state $s'$ under action $a$ is never taken, therefore $\rand{N_t (s,a,s')} = 0$ and $| \rand{N_t(s,a,s')}/ \rand{N_t(s,a)} - P^a_{s,s'} | = 0$. As a conclusion:
    \begin{equation*}
    \begin{split}
        \mathbb{P} \bigg(\norminf{\MDP - \hatMDP} > \varepsilon \bigg) & \leq \mathbb{P} \bigg( \bigcup_{s,a,s' \text{ s.t.} P^a_{s,s'} \neq 0} \bigg| \frac{\rand{N_t(s,a,s')}}{\rand{N_t(s,a)}} - P^a_{s,s'} \bigg| > \varepsilon \bigg) \\
        % & \leq \sum_{s,a,s' \text{ s.t.} P^a_{s,s'} \neq 0} \big( c_{s,a,s'} + c_{s,a} \big) \exp \big( - \varepsilon^2 \varphi_s P^a_{s,s'} t / 144 T_\varphi \big) \\
        & \leq k_{\MDP} \exp \big( - \varepsilon^2 l_{\MDP} t \big)
    \end{split}
    \end{equation*}
    by union-bound inequality, with $k_{\MDP} := \sum_{s,a,s' \text{ s.t.} P^a_{s,s'} \neq 0} ( c_{s,a,s'} + c_{s,a} )$ and $l_{\MDP} := \min_{s,a,s' \text{ s.t.} P^a_{s,s'} \neq 0} \varphi_s P^a_{s,s'} / 144 T_\varphi$. Hence the conclusion, with $l_{\MDP}, k_{\MDP}$ depending only on $\MDP$.
    %  As a conclusion:
    % \begin{equation*}
    %     t \geq \frac{-1}{\varepsilon^2 l_{\MDP}} \log \big( \delta / k_{\MDP} \big) \implies \mathbb{P} \bigg(\norminf{\MDP - \hatMDP} \leq \varepsilon \bigg) \geq 1 - \delta
    % \end{equation*}
    % where $l_{\MDP}$ does only depend on $\varphi$, so on $\MDP$.
\end{proof}

Let us now move on to the main result of this section: a sufficient condition on the learning time to have an accurate estimation of $\MDP$'s Whittle indices with high probability.

\begin{theorem}
    \label{th:precision_time}
    Let $\MDP$ be an indexable, communicating and unichain MDP. Denote by $\hatMDP_t$ the estimate of $\MDP$ at time $t$ built by $\mathBLINQ$ as in Equations (\ref{eq:approx_MDP}). Finally, let $1 \geq \delta > 0$ denote a confidence bound. \blue{Then, after a time $T_{\min} := \frac{-l_{\MDP}}{\varepsilon_{\min}^2} \log \bigg( \frac{\delta}{k_{\MDP}} \bigg)$ where $k_{\MDP}, l_{\MDP}$ come from Lemma \ref{lem:learning_time}, \EWISC~approximates the Whittle indices of $\MDP$ with probability at least $1 - \delta$.} In other words:
    \begin{equation*}
        \text{if } t \geq T_{\min}, \text{then for every state } s \in \calS, \mathbb{P} \bigg(\big| \mathEWISC (\hatMDP_t)_s - \lambda_s \big| \leq \frac{\xi_{\MDP}}{\sqrt{t}} \bigg) \geq 1 - \delta.
    \end{equation*}
    with $\xi_{\MDP} := c_{\MDP} \sqrt{\frac{-\log (\delta / k_{\MDP} )}{l_{\MDP}}}$.
\end{theorem}

\begin{remark}
    Note that $T_{\min}$ and $\xi_{\MDP}$ only depend on $\MDP$ and $\delta$.
\end{remark}

\begin{proof}
    Let us denote by $\varepsilon_\text{min} := \min \left\{ 1 / \max_{\pi \in \calA^{\calS}} \diam{P^\pi}, \frac{1}{2} \min_{\lambda_s \neq \lambda_{s'}} |\lambda_s - \lambda_{s'}| \right\}$ the minimum threshold of Assumption \ref{ass:tildeM_close}. Applying Lemma \ref{lem:learning_time} shows the existence of constants $k_{\MDP}, l_{\MDP}$ that only depend on $\MDP$ s.t.:
    \begin{equation*}
        \mathbb{P} \bigg( \norminf{\MDP - \hatMDP} \leq \sqrt{\frac{-\log (\delta / k_{\MDP} )}{l_{\MDP} t}} \bigg) \geq 1 - \delta.
    \end{equation*}
    As a consequence, if $t \geq T_{\min} := \frac{-l_{\MDP}}{\varepsilon_{\min}^2} \log \bigg( \frac{\delta}{k_{\MDP}} \bigg)$ then $\norminf{\MDP - \hatMDP_t} \leq \varepsilon_{\min}$ is satisfied with probability at least $1-\delta$. In that case, Assumption \ref{ass:tildeM_close} holds. 
    
    Moreover, let us recall that, assuming Assumptions \ref{ass:tildeM_close} hold and both MDPs share the same support, Theorem \ref{th:linear_conv} states that there exists some constant $c_{\MDP}$ that only depend on $\MDP$ which satisfies:
    \begin{equation*}
        \forall s \in \calS, \big| \mathEWISC (\hatMDP_t)_s - \lambda_s \big| \leq c_{\MDP} \norminf{\MDP - \hatMDP_t}.
    \end{equation*}

    Lastly, let us recall that from the definition of $\hatMDP_t$, it has the same support as $\MDP$. Combining everything:
    \begin{equation*}
        \text{when $t \geq T_{\min}$, for every state } s \in \calS, \mathbb{P} \bigg(\big| \mathEWISC (\hatMDP_t)_s - \lambda_s \big| \leq \frac{\xi_{\MDP}}{\sqrt{t}} \bigg) \geq 1 - \delta
    \end{equation*}
    with $\xi_{\MDP} := c_{\MDP} \sqrt{\frac{-\log (\delta / k_{\MDP} )}{l_{\MDP}}}$.
\end{proof}

%% file: sections/num_exp.tex
\blue{
    Let us now compare \BLINQ~to other learning algorithms from the existing literature. Recalling Figure \ref{fig:literature_review}, our main point of comparison is with $Q-$learning approaches: \QWI, \QWhittle, \QGI~and \WIQLUCB. We also consider an hybrid approach: \QWINN. We chose not to implement approaches based purely on NNs because of their high computational cost and sensitivity to training parameters. Every experiment involves in using the basic exploration policy of \BLINQ~(activate with probability $1 / 2$) in a simulation setting with a single arm. We focus on analyzing the Whittle indices learning speed of each algorithm. Our implementation can be found \href{https://gitlab.inria.fr/jcharles/ewisc_and_blinq}{here}\footnote{https://gitlab.inria.fr/jcharles/ewisc\_and\_blinq}

    \subsection{Learning Whittle indices for the 5-states restart problem with discounted rewards}

    In Figure \ref{fig:ex10_comparison} we compare \BLINQ, \QWI~and \QWINN~for learning the Whittle indices of the restart problem (i.e. the MDP in Figure \ref{fig:index_ex}) with a discount factor $0.9$. The $Q-$learning update parameters are as in \cite[example 4.1]{robledo_borkar2022deeplearning}.

    \begin{figure}[ht]
        \includesvg[width=0.325\textwidth]{Plots/ex10_QWI.svg}
        \includesvg[width=0.325\textwidth]{Plots/ex10_QWINN.svg}
        \includesvg[width=0.325\textwidth]{Plots/ex10_BLINQ.svg}
        \centering
        \caption{Comparison of \QWI, \QWINN~and \BLINQ, from left to right on the MDP from Figure \ref{fig:index_ex} with a discount factor $0.9$. Dotted lines represent the Whittle indices of states $0$ to $4$, from bottom to top.}
        \Description{Numerical experiments comparing our approach and $Q$-learning for learning the Whittle indices of a 5-states MDP from the literature, with discounted rewards.}
        \label{fig:ex10_comparison}
    \end{figure}

    As we can see from the leftmost plot, \QWI~struggles to approximate the Whittle indices of the least visited states ($3$ and $4$): after $30000$ iterations the distance between $\lambda_4$ (topmost dashed line) and its approximation is still significant, and does not seem stable. Generally speaking, its behavior is quite chaotic and unstable for these states. For MDPs with closer indices, the relative order of indices could change often because of this.

    \QWINN~clearly performs better than \QWI~but still converges quite slowly for state $4$. Moreover, it is slow to run as it needs to train a neural network in the meantime.

    \BLINQ~completely outperforms \QWI~and provides a more accurate and stable estimate of $\lambda_4$ than \QWINN. Overall, it has a good estimate after a few dozen iterations only.

    \subsection{Learning Whittle indices for the 5-states restart problem with average rewards}

    Let us now take the same MDP but with the average reward criterion. In this setting, \BLINQ~works out-of-the-box, but $Q-$learning needs some changes. As we can see from Figure \ref{fig:ex9_comparison}, \QWI~fails to converge. Its variant \QWhittle~works, but slowly, and it has a quite chaotic and unstable behavior, just as \QWI~in the last example.

    \begin{figure}[ht]
        \includesvg[width=0.325\textwidth]{Plots/ex9_QWI.svg}
        \includesvg[width=0.325\textwidth]{Plots/ex9_Q-Whittle.svg}
        \includesvg[width=0.325\textwidth]{Plots/ex9_BLINQ.svg}
        \centering
        \caption{Comparison of \QWI, \QWhittle~and \BLINQ, from left to right on the MDP from Figure \ref{fig:index_ex} with no discount factor. Dotted lines represent the Whittle indices of states $0$ to $4$, from bottom to top.}
        \label{fig:ex9_comparison}
        \Description{Numerical experiments comparing our approach and $Q$-learning for learning the Whittle indices of a 5-states MDP from the literature, with undiscounted rewards.}
    \end{figure}

    In this setting, comparing with \WIQLUCB~would have been relevant as it should converge with no fine-tuning required and with a $O(S)$ memory usage instead of $O(S^2)$ for \QWI~and \QWINN. However, to our surprise, we were not able to replicate their results from \cite{jonah2026WIQL}: neither by re-implementing their pseudocode, nor by running their Python code. In our simulations, \WIQLUCB~converges to the right relative order of Whittle indices (in this specific example) but not to the right values.

    \subsection{Learning Gittins indices for a 50 states MDP}

    Let us now take a MDP with state space $\calS := [|0, 49|]$, generated as in \cite[IV.A]{dhankar2025tabulardeepreinforcementlearning}. Each line of active transition matrix $(P^1_{s,:})_{a \in \calA, s \in \calS}$ is generated according to a Dirichlet distribution where each parameter is $1 / S$, and the active rewards are given by $\forall s \in \calS, r^1_s := 5+\frac{s+1}{10}$. As we are learning Gittins index, action $0$ does not induce any reward or state trantition. The discount factor is $0.9$.

    For each of the three algorithms \QWI, \QGI, \BLINQ~and at each time step $t$, Figure \ref{fig:ex8_comparison} plots the minimum, median and maximum of the absolute error vector $( | \lambda_s - \hat{\lambda}_{s, t} | )_{s \in \calS}$ where $\lambda_s$ designates the true Gittins index of state $s$ and $\hat{\lambda}_{s, t}$ designates the approximated Gittins index of state $s$ at time $t$, with experiments conducted over $50 000$ time steps. The $Q-$learning setpsizes are as in the previous example for \QWI, and as suggested in \cite{dhankar2025tabulardeepreinforcementlearning} for \QGI.

    We also provide in Table \ref{tab:ex8_comparison} a comparison of the minimum, median and maximum errors at the end of the learning process over the chosen time frame.

    \begin{figure}[ht]
        \includesvg[width=0.325\textwidth]{Plots/ex11_QWI.svg}
        \includesvg[width=0.325\textwidth]{Plots/ex11_QGI.svg}
        \includesvg[width=0.325\textwidth]{Plots/ex11_BLINQ.svg}
        \centering
        \caption{Comparison of the minimum, median and maximum error as a function of the time $t$ for \QWI, \QGI~and \BLINQ~on the MDP from \cite[section IV.A]{dhankar2025tabulardeepreinforcementlearning} with a discount factor $0.9$. Experiments are conducted over $50 000$ time steps. Note that \BLINQ~waits for every state-action pair to be explored before running \EWISC.}
        \label{fig:ex8_comparison}
        \Description{Numerical experiments comparing our approach and $Q$-learning for learning the Gittins indices of a many-states MDP from the literature.}
    \end{figure}

    \begin{table}
        \begin{center}
            \begin{tabular}{ | l | c | c | c | }
            \hline
            Algorithm & \QWI & \QGI & \BLINQ \\ \hline
            Maximum error at last time step & $10^{12}$ & 2.808 & 0.135 \\ \hline
            Median error at last time step & 6.640 & 0.012 & 0.002 \\ \hline
            Minimum error at last time step & 0.008 & 0.0 & 0.0 \\ \hline
            \end{tabular}
            \caption{Comparison of the errors of the three algorithms after $50 000$ time steps, rounded to 3 decimal places. Note that $Q-$learning implementations may suffer from floating-point arithmetic errors.}
            \label{tab:ex8_comparison}
        \end{center}
    \end{table}

    It ls clear that \BLINQ~significantly outperforms both \QGI~and \QWI~when it comes to the maximum error, and \QWI~does not produce accurate estimates of indices after $50 000$ learning steps.
}

\subsection{Time and space complexity}

Both time and space complexity of each algorithm are summarized in Table \ref{tab:time_space__comparison}.

At each time step \BLINQ~requires a time complexity of $O(1)$ to update the quantities $N_t(\cdot)$ defined in (\ref{eq:approx_MDP}). When it runs \EWISC, it needs a time complexity $O(S^2)$ to compute the estimated transition matrix $\hat{P}$ plus $O(K S^2)$ where $K$ is the number of BO policies computed, which is $S+1$ if the approximated MDP $\hatMDP$ is indexable. An efficient optimization consists in increasing the number of time steps between each consecutive runs of \EWISC, e.g. by doubling it. This allows the computational cost of \EWISC~to be negligible compared to the cost of update, in the long run. During our experiments, estimated MDPs were almost always indexable, hence a complexity of $O(S^3)$ for each run of \EWISC.

In contrast, \QWhittle, \QWI~and \QGI~have a runtime that does not depend on the time step $t$. They work differently than \BLINQ~in the sense that the learning can be decoupled for each state, in case the user does not need to approximate all Whittle indices. As a result, the computational complexity can be reduced: they require a time complexity of $O(1)$ per state at each time step, so $O(S)$ when learning all indices. They do not require any additional time complexity.

In our tests with \BLINQ, we multiplied the number of time steps between consecutive runs of \EWISC~by some constant greater than $1$. In the long run the total execution time of \EWISC~and \WISC~by \BLINQ~was orders of magnitude smaller than the accumulated update time needed by \QWhittle, \QWI~and \QGI, as expected.

In terms of memory usage, \BLINQ~requires to compute transition matrices and therefore needs a space complexity of $O(S^2)$. \QWhittle, \QWI~and \QGI~work differently as they need a table of size $O(S)$ to learn each Whittle or Gittins index the decision maker wants to learn. However, the overall complexity if they want to learn all state indices at once is still $O(S^2)$.

To summarize, on the one hand, the $Q$-learning based approaches can be decoupled on a state-by-state scheme. On the other hand, \BLINQ~is more time-efficient in the long run (after a few hundreds time steps in our testing) than its $Q$-learning counterparts, and needs the same amount of memory when estimating the indices of all states.  In addition, \QWhittle, \QWI~and \QGI~---as usual with $Q$-learning based approaches--- are very sensitive to update parameters. The number of steps needed to get a good approximation of Whittle indices can be multiplied by several hundreds depending on the user-chosen stepsizes. In contrast, \BLINQ~is parameter-free, thus it does not require the fine-tuning of any parameter. What is more, we provided a detailed analysis of its speed of convergence in $O(1 / \sqrt{T})$.

\begin{table}
    \begin{center}
        \begin{tabular}{ | l | c | c | c | c | }
          \hline
          Algorithm & \QWhittle & \QWI & \QGI & \BLINQ \\ \hline
          Time complexity over $T$ time steps & $O(S . T)$ & $O(S . T)$ & $O(S . T)$ & \begin{tabular}{@{}c@{}} $O(T + S^3 . f(T))$ \\ $f(T)$ is the number of runs of \EWISC. \\ In our experiments, $f(T) = \log{T}$. \end{tabular} \\ \hline
          Space complexity & $O(S^2)$ & $O(S^2)$ & $O(S^2)$ & $O(S^2)$ \\ \hline
          Requires fine-tuned parameters & Yes & Yes & Yes & No \\ \hline
        \end{tabular}
        \caption{Comparison of the total time and space complexity of each algorithm to learn all $S$ Whittle indices of $\MDP$.}
        \label{tab:time_space__comparison}
    \end{center}
\end{table}

%% file: sections/appendix.tex
\section{Proof of Theorem \ref{th:distinct_WI}: interior points of the subset of indexable MDPs}
\label{app:sec:proof_distinct_WI}

\begin{proof}
    Similarly to most of this paper's results, this proof relies on the geometric analysis of the Whittle index introduced in \cite{gast2023testing} and summarized in \textsection \ref{par:charact_of_WI}. We first choose a $\delta > 0$ such that for any state $s \in \calS$, the optimal activation advantage function of state $s$ in $\hatMDP$ (named $\hat{\alpha}^*_s$ and defined in Equation (\ref{eq:act_adv_optimal})) cancels out only locally around $\lambda_s$: the Whittle index of state $s$ in $\MDP$. We then characterize the variations of $\hat{\alpha}^*_s$ locally around $\lambda_s$. Combining the former and the latter result allows us to deduce that $\hat{\alpha}^*_s$ crosses the x-axis only once on $\mathbb{R}$, thus making $\hatMDP$ indexable.

    Let $(\lambda_s)_{s \in \calS}$ be the Whittle indices of $\MDP$, and let $\hatMDP:=\big(\calS, \calA, (\hat{P}^a)_{a \in \calA}, (r(a,\cdot))_{a \in \calA}\big)$ be a MDP that has the same support as $\MDP$, as defined in Equation (\ref{eq:same_support}); and the same rewards as well. Without loss of generality let us number states so that $\lambda_0 < \cdots < \lambda_{S-1}$, and for convenience let us define $\lambda_{-1} := - \infty$ and $\lambda_S := + \infty$. 
    
    Finally, let us define $\varepsilon_\text{min} := \min \left\{ 1 / \max_{\pi \in \calA^{\calS}} \diam{P^\pi}, \frac{1}{2} \min_{0 \leq s < S - 1} |\lambda_s - \lambda_{s+1}| \right\}$, then Assumption \ref{ass:tildeM_close} is satisfied whenever $\norminf{\MDP - \hatMDP}\le\varepsilon_{\min}$.

    Suppose that $\norminf{\MDP - \hatMDP}\le\varepsilon_{\min}$, so that Assumption \ref{ass:tildeM_close} holds. In addition, $\MDP$ is unichain and indexable, and has the same support as $\hatMDP$, therefore we apply Lemma \ref{lem:proof_th_lin_conv}: there exists constants $\delta_{\MDP}, c_{\MDP}$ that depend only on $\MDP$ such that for any state $s \in \calS$:
    \begin{enumerate}[label=(\roman*)]
        \item if $\norminf{\hatMDP - \MDP} \leq \delta_{\MDP}$ then for any $\lambda \in [\lambda_s + \delta_{\MDP}, \lambda_{s+1} - \delta_{\MDP}]$ the BO policy in the $\lambda-$ penalized MDP $\hatMDP(\lambda)$ is unique and is the same as in $\MDP(\lambda)$, \label{enum:fact1} 
        \item there exists a penalty $\hat{\lambda} \in \mathbb{R}$ such that such that $\hat{\alpha}^*_s(\hat{\lambda}) = 0$, and all such penalties satisfy $| \lambda_s - \hat{\lambda} | \leq c_{\MDP} \norminf{\MDP - \hatMDP}.$ \label{enum:fact2}
    \end{enumerate}

    Let $\delta > 0$ be a real number that is strictly lower than $\varepsilon_{\min}$, $\delta_{\MDP}$ and $\frac{1}{2 c_{\MDP}} \min_{0 \leq s < S-2} | \lambda_s - \lambda_{s+1} |$, and suppose that $\norminf{\MDP - \hatMDP} \leq \delta$. Note that $c_{\MDP} \norminf{\MDP - \hatMDP} < c_{\MDP} \delta \leq \frac{1}{2} \min_{0 \leq s < S-2} | \lambda_s - \lambda_{s+1} |$ Recall, our goal is to prove that there exists exactly one penalty value $\hat{\lambda_s} \in \mathbb{R}$ such that $\hat{\alpha}^*_s(\hat{\lambda_s}) = 0$. First, by direct reformulation of \ref{enum:fact1}:
    \begin{enumerate}
        \item for $\lambda_{s-1} + \delta < \lambda < \lambda_s - \delta$, the unique BO policy in $\hatMDP(\lambda)$ is the same as in $\MDP(\lambda)$, named $\pi_s$. This implies $\hat{\alpha}_s^* (\lambda) = \hat{\alpha}_s^{\pi_s}(\lambda) > 0$. \label{enum:step1}
        \item Similarly, for $\lambda_{s} + \delta < \lambda < \lambda_{s+1} - \delta$, the unique BO policy in $\hatMDP(\lambda)$ is the same as in $\MDP(\lambda)$, named $\pi_{s+1}$, implying $\hat{\alpha}_s^* (\lambda) = \hat{\alpha}_s^{\pi_{s+1}}(\lambda) < 0$. \label{enum:step2}
    \end{enumerate}
    These two observations allow us to deduce the following facts which shall conclude our proof:
    \begin{enumerate}
    \setcounter{enumi}{2}
        \item for any $\lambda \in [ \lambda_s - \delta, \lambda_s + \delta ]$, the BO policies in $\hatMDP(\lambda)$ are either $\pi_s$ or $\pi_{s+1}$ (or both). This implies $\hat{\alpha}_s^* (\lambda) = \hat{\alpha}_s^{\pi_s}(\lambda)$ or $\hat{\alpha}_s^* (\lambda) = \hat{\alpha}_s^{\pi_{s+1}}(\lambda)$. \label{enum:step3}
        \item Finally, both $\hat{\alpha}_s^{\pi_{s}}$ and $\hat{\alpha}_s^{\pi_{s+1}}$ are strictly nonincreasing on $\mathbb{R}$. \label{enum:step4}
    \end{enumerate}

    % \begin{figure}[ht]
    %     \input{IPE_figures/proof_distinctWI.tex}
    %     \centering
    %     \caption{Summary of facts (\ref{enum:step1}) to (\ref{enum:step4})---indicated in red---and their relative roles.}
    %     \label{fig:proof_distinctWI}
    %     \Description{Figure summarizing the proof according to which a MDP with distinct Whittle indices has an indexable neighborhood.}
    % \end{figure}

    Indeed, from (\ref{enum:step3}) and (\ref{enum:step4}) $\hat{\alpha}_s^*$ is strictly nonincreasing on $[\lambda_s - \delta, \lambda_s + \delta]$. From \ref{enum:fact2}, it crosses the x-axis at least once on this interval, and cannot do so elsewhere. As a result, it cancels out exactly once on $\mathbb{R}$, thus $\hatMDP$ is indexable. What is left is to prove facts (\ref{enum:step3}) and (\ref{enum:step4}).

    \emph{Proof of (\ref{enum:step3}): } let $s' \neq s$ be any other state, because of \ref{enum:fact2} the function $\hat{\alpha}^*_s : \mathbb{R} \rightarrow \mathbb{R}$ can cancel out only at a distance from $\lambda_s$ lower than $c_{\MDP} \norminf{\MDP - \hatMDP} \leq c_{\MDP} \delta < \frac{1}{2} \min_{0 \leq s < S - 1} |\lambda_s - \lambda_{s+1}|$. As a consequence, the function $\hat{\alpha}_{s'}^*$ does not cancel out on $[\lambda_s - \delta, \lambda_s + \delta]$. As it is continuous, its sign does not change on this interval, therefore by recalling Equation (\ref{eq:charact_BO}) one of the two following facts is true:
    \begin{subequations}
    \begin{align}
        \forall \lambda \in [\lambda_s - \delta, \lambda_s + \delta], \hat{\alpha}_{s'}^* (\lambda) > & 0\text{: all BO policies in $\hatMDP(\lambda)$ contain $s'$, or} \label{eq:all_BO_contain} \\
        \forall \lambda \in [\lambda_s - \delta, \lambda_s + \delta], \hat{\alpha}_{s'}^* (\lambda) < & 0\text{: no BO policies in $\hatMDP(\lambda)$ contain } s'. \nonumber 
    \end{align}
    \end{subequations}
    As a consequence, given a $\lambda \in [\lambda_s - \delta, \lambda_s + \delta]$ there exists at most two BO policies in $\hatMDP(\lambda)$: the one that contains all states $s' \neq s$ satisfying Equation (\ref{eq:all_BO_contain}), as well as $s$, and the one that contains the same states except for $s$. Now, let us recall that $\pi_s$ is BO in $\hatMDP(\lambda_s - \delta)$, and $\pi_{s+1}$ is BO in $\hatMDP(\lambda_s + \delta)$. As a conclusion, in $\hatMDP(\lambda)$, only $\pi_s$ or $\pi_{s+1}$ can be BO.

    \emph{Proof of (\ref{enum:step4}): } let $\hat{\lambda}_s^\text{first} := \inf \{\lambda \in [\lambda_s - \delta, \lambda_s + \delta] ; \hat{\alpha}_{s}^* (\lambda) = 0 \}$ and $\hat{\lambda}_s^\text{last} := \sup \{\lambda \in [\lambda_s - \delta, \lambda_s + \delta] ; \hat{\alpha}_{s}^* (\lambda) = 0 \}$ be the first and last vanishing points of $\hat{\alpha}_s^*$, respectively. On the interval $\lambda \in [\lambda_{s-1} + \delta, \hat{\lambda}_s^\text{first})$, we have $\hat{\alpha}_s^* (\lambda) > 0$. As the sign of all optimal activation advantages of any other state also stays constant on that interval, the BO policy in $\hatMDP(\lambda)$ is $\pi_s$, and the optimal activation advantage is $\hat{\alpha}_s^{\pi_s}$, which is affine. Because it vanishes in $\hat{\lambda}_s^\text{first}$, its slope is strictly negative. We prove in a completely analogous way that the slope of $\hat{\alpha}_s^{\pi_{s+1}}$ is strictly negative too. 

    Let us consider $\hat{\alpha}_s^*$ on the interval $[\lambda - \delta, \lambda + \delta]$, from fact (3) it is pointwise either equal to $\hat{\alpha}_s^{\pi_s}$ or $\hat{\alpha}_s^{\pi_{s+1}}$, in both cases it is strictly nonincreasing therefore
$\hat{\alpha}_s^*$ is strictly nonincreasing on $[\lambda - \delta, \lambda + \delta]$.
 \end{proof}

\section{Low-level desription and time complexity analysis of Algorithm \ref{alg:EWISC}}
\label{app:sec:EWISC}

\subsection{Pseudocode}

\begin{algorithm}[htbp]
    \caption{\textbf{E}xtended \textbf{WISC}}
    \label{alg:EWISC}
    \begin{algorithmic}
    \State \textbf{Input:} a MDP $\hatMDP$
    \State \textbf{Process:} \EWISC~computes indices $- \infty = \mu_0 \leq \dots \leq \mu_{K+1} = + \infty$ and BO policies $\calS = \pi_0, \dots, \pi_K = \emptyset$ as described in \textsection\ref{par:charact_of_EWISC}
    \State \textbf{Output:} for each $s \in \calS$, the average of all penalties $\mu \in \mathbb{R}$ s.t. $\alpha_s^*(\mu) = 0$.
    \State $\mu_0 \gets -\infty, \pi_0 \gets \calS$, $i \gets 0$
    \State $\mathtt{buff} \gets \emptyset$ \Comment{Initialise buffer}
    \For{each state $s \in \calS$}
        \State $\mathtt{intersections[s]} \gets \emptyset$ \Comment{\texttt{intersections[s]} will contain roots of $\alpha_s^*$}
    \EndFor
    % \State For each state $s \in \calS, \mathtt{intersections[s]} \gets \{\emptyset\}$ \Comment{\texttt{intersections[s]} will contain the penalties at which $\alpha_s^*$ cancels out}
    \While{$\pi_i \neq \emptyset$}
        \State $\mathtt{next\_intersection} \gets \{\min \{ \lambda \geq \mu_i | \alpha_s^{\pi_i} (\lambda) = 0 \} ; s \in \calS \setminus \mathtt{buff}\}$ \Comment{\texttt{next\_intersection[s]} is the next intersection of $\alpha_s^{\pi_i}$ with the x-axis}
        % \State For each state $s \in \calS \setminus \mathtt{buff}, \mathtt{next\_intersection[s]} \gets \min \{ \lambda \geq \mu_i | \alpha_s^{\pi_i} (\lambda) = 0 \}$
        % \For{each state $s \in \calS \setminus \mathtt{buff}$}
        %     \State $\mathtt{next\_intersection[s]} \gets \{ \lambda \geq \mu_i | \alpha_s^{\pi_i} (\lambda) = 0 \}$ % \Comment{\texttt{next\_intersection[s] is the next intersection of $\alpha_s^{\pi_i}}$ with the x-axis}
        % \EndFor
        \State $\sigma \gets \argmin_{s \in \calS \setminus \mathtt{buff}} \mathtt{next\_intersection[s]}$
        \State $\mu \gets \mathtt{next\_intersection[\sigma]}$

        \If{$\mu = \mu_i$}
            \State $\mathtt{buff} \gets \mathtt{buff} \cup \{ \sigma \}$ \Comment{Add $\sigma$ to the buffer}
        \Else
            \State $\mathtt{buff} \gets \{ \sigma \}$ \Comment{Reset buffer}
        \EndIf
        \State $\mathtt{intersections[\sigma]} \gets \mathtt{intersections[\sigma]} \cup \{\mu\}$ \Comment{Update intersections table}
        \State $\mu_{i+1} \gets \mu$ \Comment{Update indices sequence}
        \State $\pi_{i+1} \gets \pi_i \Delta \{ \sigma \}$ \Comment{By definition, $A \Delta B := (A \cup B) \setminus (A \cap B)$}

        \State{$i \gets i+1$}
    \EndWhile \\
    \Return the vector $\{\mathtt{average(intersections[s])} ; s \in \calS\}$
    \end{algorithmic}
\end{algorithm}

Let us now explain how \EWISC~ works. It proceeds by constructing the sequence $(\mu_i)_i$ in a increasing order. It starts with a penalty $\lambda =  \mu_0 = - \infty$, for which the policy $\pi_0 = \calS$ is necessarily BO in $\MDP(\lambda)$. Starting from a policy $\pi_i$, \EWISC~ then finds the smallest $\lambda$ (called $\mu_{i+1}$) for which adding or removing a state from $\pi_i$ creates another BO policy in $\MDP(\lambda)$.  When $\lambda$ becomes so big that $\pi = \emptyset$ is BO in $\MDP(\lambda)$, it stops. 

% When compared to \WISC, \EWISC~ can add or remove states from $\pi_i$ whereas \WISC~ only considers taking states out of $\pi$ while $\lambda$ grows, and returns \texttt{"Not indexable"} if it finds a contradiction. In contrast, \EWISC~also considers adding states back to $\pi$, thus computing a BO policy sequence even when $\MDP$ is not indexable.

In details, \EWISC~starts with $i:=0, \mu_i := - \infty, \pi_i := \calS$ and computes the activation advantage vector under $\pi_i$, that is $(\alpha_s^{\pi_i} (\lambda))_{s \in \calS}$ defined in Definition \ref{def:act_advantage}. From \cite{gast2023testing} each component is affine in $\lambda$, so \EWISC~computes the state whose activation advantage under $\pi_i$ cancels out the soonest: $\sigma := \argmin_{s \in \calS} \big\{ \min \{ \mu \geq \mu_i | \alpha^{\pi_i}_s (\mu) = 0 \} \big\}$. Then, it updates $\mu_i$ accordingly: $\mu_{i+1} := \min \{ \mu \geq \mu_i | \alpha^{\pi_i}_\sigma (\mu) = 0 \}.$

\EWISC~then updates the BO policy by setting $\pi_{i+1} := \pi_i \cup \{\sigma\}$ if $\sigma$ is not already in $\pi_i$, or $\pi_{i+1} := \pi_i \setminus \{\sigma\}$ otherwise. In other words, $\pi_{i+1} := \pi_i \Delta \{ \sigma \}$ where $A \Delta B := (A \cup B) \setminus (A \cap B)$ denotes the symmetric difference between two sets. Finally, $i$ is incremented and the process is repeated until $\pi_i = \emptyset$. 

When successive penalties are identical, the set \texttt{buff} contains the states that have been considered by \EWISC~for that penalty. At step $j$, assuming the considered penalty is $\mu_j$ since step $i < j$ (in other words $\mu_{i-1} < \mu_i = \mu_{i+1} = \dots = \mu_j$) we have 
\begin{equation*}
    \texttt{buff} = \bigcup_{k = i}^j \{ \sigma \text{ is the state considered at step $k$ by \EWISC's \texttt{while} loop} \}.
\end{equation*}

\subsection{Time complexity} 

Each pass in the \texttt{while} loop requires the computation of all affine functions $\alpha_s^{\pi_i}$ for all $s \in \calS$. As explained in \cite[Section 3.4, Equation 11]{gast2023testing}, this can be done naively by some matrix multiplications and inversions. Indeed, representing the activation advantages as a column vector $\alpha^{\pi_i} (\lambda) := \big( \alpha^{\pi_i}_s (\lambda) | s \in \calS \big)^\top$, the following holds:
\begin{equation}
    \alpha^{\pi_i} (\lambda) = r^1 - r^0 - \lambda \mathbf{1}^\top + \big(P^1 - P^0 \big) (A^{\pi_i})^{-1} (r^{\pi_i} - \lambda \texttt{bin}^{\pi_i} ),\label{eq:advantage_computation}
\end{equation}
where:
\begin{equation*}
    \begin{split}
        A^{\pi_i} & := 
        \begin{bmatrix}
            1 & & \\
            1 & \ddots & \\
            1 & & 1 \\
        \end{bmatrix} - 
        \begin{bmatrix}
            0 & P(\pi_i(0))_{0,1} & \dots & P(\pi_i(0))_{0,S-1} \\
            \vdots  & \vdots & & \vdots \\
            0 & P(\pi_i(S-1))_{S-1,1} & \dots & P(\pi_i(S-1))_{S-1,S-1} \\
        \end{bmatrix},
        \\
        \texttt{bin}^{\pi_i} & := [ \mathds{1}_{s \in \pi} | s \in \calS ]^\top, \\
    \end{split}
\end{equation*}
and $A^{\pi_i}$ is invertible because $\MDP$ is unichain \cite{puterman2014markov}.

The time complexity of the first pass in the \texttt{while} loop is $O(S^3)$ because of the matrix inversion $(A^{\pi_0})^{-1}$. However, as in \cite[Section 4.2]{gast2023testing} the computation of $(A^{\pi_{i+1}})^{-1}$ can be done efficiently from $(A^{\pi_i})^{-1}$ in $O(S^2)$ by using the Sherman-Morrison formula.

Indeed, we only add or remove one state at a time from $\pi_i$ to get $\pi_{i+1}$, so the former can be written as $\pi_{i+1} = \pi_i + e_i$ where $e_i$ is a vector having a $1$ or $-1$ at only one coordinate and $0$'s elsewhere. This implies that the matrix $A^{\pi_{i+1}}$ has a one-dimensional change compared to $A^{\pi_{i}}$: it can be written as $A^{\pi_{i+1}} = A^{\pi_{i}} + p q^\top$ for some vectors $p, q \in \mathbb{R}^S$. Finally, the Sherman-Morrison formula provides an way to compute $(A^{\pi_{i+1}} + p q^\top)^{-1}$ by only doing a constant number of multiplications with $A^{\pi_{i}}, p, q$:

\begin{equation*}
    \big( A + p q^\top \big)^{-1} = A^{-1} + \frac{A^{-1} p q^\top A^{-1}}{1 + q^\top A^{-1} p}.
\end{equation*}

All in all, this implementation allows for the following complexity: the first computation of $A^{\pi_0}$ is done in $O(S^3)$, next passes in the \texttt{while} loop are done in $O(S^2)$. As a conclusion, the time complexity of \EWISC~is $O(K S^2)$ where $K \geq S$ is the number of BO policies computed by \EWISC.

\section{Proof of Lemma \ref{lem:proof_th_lin_conv}: convergence rate of \EWISC's output as a function of its input}
\label{app:sec:proof_lem2}

This is the proof of the main result of our paper, and is quite technical. As explained earlier in \textsection\ref{par:sketch_proof_lem2}, we shall use five lemmas as intermediary results before moving on to the proof itself. 

Let us recall that $\hatMDP:=\big(\calS, \calA, (\hat{P}^a)_{a \in \calA}, (r^a)_{a \in \calA}\big)$ denotes a MDP whose rewards are identical to the ones of $\MDP$. 

% \begin{figure}[ht]
%     \input{IPE_figures/proof_lem2_new.tex}
%     \centering
%     \caption{Summary of the roles of our intermediate lemmas in the proof of Lemma \ref{lem:proof_th_lin_conv}, for uniformly bounding the optimal advantage gap for any ``relevant'' penalty $\lambda$:}
%     \label{fig:proof_th_linear_convergence}
%         \Description{Figure summarizing the proof of the main result.}
% \end{figure}

\subsection{Lemma \ref{lem:WI_bounded}: an explicit bound on the possible Whittle index values}

We recall the definition of a Markov chain's diameter: if $P$ is the transition kernel of a unichain Markov chain with recurrent component $\calS_r$, its diameter $\diam{P}$ is defined as the maximum time needed to go from one state to a state in its unique recurrent class: denoting by $\rand{\tau_{s,s'}}$ the random variable representing the number of time steps necessary to go from state $s$ to state $s'$:
\begin{equation*}
    \diam{P} := \max_{s \in \calS,s' \in \calS_r} \mathbb{E}^P [\rand{\tau_{s,s'}}].
\end{equation*}

\begin{lemma}
    \label{lem:WI_bounded}
    Assume that both policies $\calS$ and $\emptyset$ are unichain, so that one can consider the diameters of their induced Markov chains. The two following facts hold: 
    \begin{itemize}
        \item if $\lambda \leq - \norminf{r^1 - r^0} - \frac{1}{2} \range{r^1} \diam{P^1} \norminf{P^1 - P^0}$ then $\calS$ is BO in $\MDP(\lambda)$,
        \item if $\lambda \geq \norminf{r^1 - r^0} + \frac{1}{2} \range{r^0} \diam{P^0} \norminf{P^1 - P^0}$ then $\emptyset$ is BO in $\MDP(\lambda)$.
    \end{itemize}
\end{lemma}

\begin{proof}
    Let $\pi := \calS$ the policy that chooses action $1$ in all states, and $\pi' := \emptyset$ the policy that chooses action $0$ in all states. Let $\lambda \in \mathbb{R}$ be a penalty, and to ease the notation define $\Delta_r := \norminf{r^1 - r^0}$ and $\Delta_P := \norminf{P^1 - P^0}$. 

    Before moving on to the proof, we work out a small deviation lemma. If $p,q \in \mathbb{R}^S$ are $S$-dimensional probability vectors and $b$ is a $S$-dimensional vector:
    \begin{equation}
        \label{eq:first_dev_lemma}
        |(p-q) \cdot h| \leq \frac{1}{2} \range{b} \normone{p-q},
    \end{equation}
    where $\range{b} := \max_{s \in \calS} b_s - \min_{s \in \calS} b_s$ designates the \emph{span} of vector $b$, and $\normone{\cdot}$ is the $1-$norm for vectors: $\normone{p} := \sum_s | p_s |$.

    \emph{Proof of the first bullet point:} suppose $\lambda \leq - \Delta_r - \frac{1}{2} \range{r^1} \diam{P^1} \Delta_P$ and let $s \in \calS$ be a state. Then:
    \begin{equation*}
    \begin{split}
        \alpha_s^\pi (\lambda) & = r^1_s - \lambda - r^0_s + (P^1_{s,\cdot} - P^0_{s,\cdot}) \cdot b^\pi (\lambda) \\
        & \geq - \norminf{r^1 - r^0} - \frac{1}{2} \range{b^\pi (\lambda)} \normone{P^1_{s,\cdot} - P^0_{s,\cdot}} - \lambda \\
        & \geq - \norminf{r^1 - r^0} - \frac{1}{2} \range{r^\pi (\lambda)} \diam{P^\pi} \norminf{P^1 - P^0} - \lambda 
    \end{split}
    \end{equation*}
    where the last step comes from \cite[Lemma D.3]{boone2025asymptoticallyoptimalregretcommunicating}. Here, $r^\pi (\lambda) = r^1 - \lambda \cdot 1^\top$ so $\range{r^\pi (\lambda)}$ does not depend on $\lambda$. Moreover, $P^\pi = P^1$, and so:
    \begin{equation*}
    \begin{split}
        \alpha_s^\pi (\lambda) & \geq - \norminf{r^1 - r^0} - \frac{1}{2} \range{r^1} \diam{P^1} \norminf{P^1 - P^0} - \lambda \geq 0.
    \end{split}
    \end{equation*}
    As a consequence, $\pi$ is a BO policy in $\MDP(\lambda)$ for $\lambda \leq - \Delta_r - \frac{1}{2} \range{r^1} \diam{P^1} \Delta_P$.

    \emph{Proof of the second bullet point:} like before, suppose $\lambda \geq \Delta_r + \frac{1}{2} \range{r^0} \diam{P^0} \Delta_P$ and let $s \in \calS$ be a state. Then:
    \begin{equation*}
    \begin{split}
        \alpha_s^{\pi'} (\lambda) & \leq \norminf{r^1 - r^0} + \frac{1}{2} \range{r^{\pi'} (\lambda)} \diam{P^{\pi'}} \norminf{P^1 - P^0} - \lambda \leq 0.
    \end{split}
    \end{equation*}
    As a consequence, $\pi'$ is a BO policy in $\MDP(\lambda)$.    
\end{proof}

\begin{corollary}
    \label{cor:WI_bounded}
    If $\MDP$ is indexable, all its Whittle indices $(\lambda_s)_{s \in \calS}$ satisfy:
    \begin{equation*}
        - \norminf{r^1 - r^0} - \frac{1}{2} \range{r^1} \diam{P^1} \norminf{P^1 - P^0} \leq \lambda_s \leq \norminf{r^1 - r^0} + \frac{1}{2} \range{r^0} \diam{P^0} \norminf{P^1 - P^0}.
    \end{equation*}
\end{corollary}

\subsection{Lemma \ref{lem:conv_of_adv}: Lipschitz continuity of the activation advantage for a fixed state-policy pair}

Let us consider a state $s$ and a policy $\pi$ and let us upper bound the activation advantage gap of $s$ under policy $\pi$ when considered in $\MDP$ on one hand and $\hatMDP$ on the other hand, in terms of the distance between both MDPs.

\begin{lemma}
    \label{lem:conv_of_adv}
    Suppose that $\MDP$ is unichain and indexable, that $\hatMDP$ is a sufficiently accurate estimate of $\MDP$ as in Assumption \ref{ass:tildeM_close} and finally that both MDPs have the same support (see Equation (\ref{eq:same_support})). There exists some $c_\alpha > 0$ that only depends on $\MDP$ s.t. for every state $s \in \calS$ and policy $\pi \in \calA^{\calS}$, denoting by $\alpha_s^\pi$ and $\hat{\alpha}_s^\pi$ the activation advantages of $s$ under policy $\pi$ in $\MDP$ and $\hatMDP$, respectively:
    \begin{equation*}
        | \alpha_s^\pi - \hat{\alpha}_s^\pi | \leq c_\alpha \norminf{\MDP - \hatMDP}.
    \end{equation*}
\end{lemma}

\begin{proof}
    First, we deduce from the small deviation lemma introduced in Equation (\ref{eq:first_dev_lemma}) the following: if $p,q \in \mathbb{R}^S$ are $S$-dimensional probability vectors and $h, \hat{h} \in \mathbb{R}^S$ are two $S$-dimensional vectors:
    \begin{equation}
        \label{eq:second_dev_lemma}
        |p \cdot h - q \cdot \hat{h}| \leq \frac{1}{2} \range{h} \normone{p-q} + \norminf{h - \hat{h}}.
    \end{equation}
    Then:
    \begin{equation*}
    \begin{split}
        | \alpha_s^\pi - \hat{\alpha}_s^{\pi} | & = \big| (P^1_{s,\cdot} - P^0_{s,\cdot}) \cdot b^\pi + (\hat{P}^1_{s,\cdot} - \hat{P}^0_{s,\cdot}) \cdot \hat{h}^\pi \big| \text{ by definition} \\
        & \leq \frac{1}{2} \range{b^\pi} \normone{P^1_{s,\cdot} - \hat{P}^1_{s,\cdot}} + \frac{1}{2} \range{b^\pi} \normone{P^0_{s,\cdot} - \hat{P}^0_{s,\cdot}} + 2 \norminf{b^\pi - \hat{h}^\pi} \\
        & \leq \frac{1}{2} \range{b^\pi} \norminf{P^1 - \hat{P}^1} + \frac{1}{2} \range{b^\pi} \norminf{P^0 - \hat{P}^0} + 2 \norminf{b^\pi - \hat{h}^\pi} \\
        & \leq \range{b^\pi} \norminf{\MDP - \hatMDP} + 2 \norminf{b^\pi - \hat{h}^\pi}.
    \end{split}
    \end{equation*}

    Next, we use a result from \cite[Lemma D.10]{boone2025asymptoticallyoptimalregretcommunicating}, which we can apply because $\MDP$ is unichain, indexable and has the same support as $\hatMDP$:
    \begin{equation*}
        \norminf{b^\pi - \hat{h}^\pi} \leq \big( 2 \diam{\hat{P}^{\pi}} \range{b^\pi} + \frac{1}{2} \range{b^{\pi, 1}} \big) \norminf{P^\pi - \hat{P}^\pi},
    \end{equation*}
    where $\range{b^{\pi,1}}$ designates the $1$-order bias\footnote{Recall that the bias $b^{\pi}$---also called the $0$-order bias---is the solution of the equation $g^\pi+b^\pi=r^\pi+P^\pi b^\pi$. The $1$-order bias $b^{\pi,1}$ is the solution of the equation $b^{\pi,1}=b^\pi+P^\pi b^{\pi,1}$. When $P^\pi$ is unichain, it is defined up to an additive constant---similarly to $b^\pi$---which implies that $\range{b^{\pi,1}}$ is uniquely defined. } under policy $\pi$. Then, from \cite[Equation D.12]{boone2025asymptoticallyoptimalregretcommunicating} with Assumption~\ref{ass:tildeM_close}: $\diam{\hat{P}^\pi} \leq 2 \diam{P^\pi}$. Combining all equations yields the desired bound with:
    \begin{equation*}
        c_\alpha := \max_{\pi \in \calA^{\calS}} \big\{ \range{b^\pi} + 8 \diam{P^\pi} \range{b^\pi} + \frac{1}{2} \range{b^{\pi,1}} \big\}.
    \end{equation*}
\end{proof}

Let $\lambda \in \mathbb{R}$ be a parameter, $\pi \in \calA^\calS$ be any policy, and let us consider the $\lambda-$penalized MDP $\MDP (\lambda)$. As proved in \cite{gast2023testing}, its bias vector under policy $\pi$ noted $b^\pi(\lambda)$ is componentwise affine in $\lambda$. As a result, its span $\range{b^\pi(\lambda)}$ is also affine in $\lambda$. Similarly, $\range{b^{\pi, 1}(\lambda)}$ is affine in $\lambda$ too.

Subsequently, both $\range{b^\pi(\cdot)}$ and $\range{b^{\pi, 1}(\cdot)}$ are uniformly bounded on any closed real interval $[a,b]$. As a consequence, the following holds from Lemma \ref{lem:conv_of_adv}

\begin{corollary}
    \label{cor:WI_unif_cv_of_adv}
    For every real interval $[a,b] \subset \mathbb{R}$, there exists a constant $c_{\lambda}$ that depends only on $\MDP, a \text{ and } b$ s.t. for every state $s \in \calS$ and policy $\pi \in \calA^\calS$:
    \begin{equation*}
        \max_{\lambda \in [a,b]} | \alpha_s^\pi (\lambda) - \hat{\alpha}_s^\pi (\lambda) | \leq c_{\lambda} \norminf{\MDP - \hatMDP}
    \end{equation*}
\end{corollary}

\subsection{Lemma \ref{lem:uniqueBO_close}: BO policy in MDPs that are sufficiently close}

The following lemma formalizes the fact that if a policy is the only BO policy in $\MDP$, then it is also the only BO policy in $\hatMDP$, provided it is sufficiently close to $\MDP$. Together with the previous Lemma \ref{lem:conv_of_adv}, this result shall allow us to bound the optimal activation advantage gap in terms of the distance between $\MDP$ and $\hatMDP$.

\begin{lemma}
    \label{lem:uniqueBO_close}
    Suppose that $\MDP$ is unichain, that $\hatMDP$ is an accurate estimate of it as in Assumption $\ref{ass:tildeM_close}$ and finally that both MDPs have the same support as defined in Equation (\ref{eq:same_support}). In addition, let us assume that there exists a unique BO policy $\pi^* \in \calA^\calS$ in $\MDP$. Then there exists a threshold $\delta_0 > 0$ s.t.
    \begin{equation*}
        \text{if } \norminf{\MDP - \hatMDP} \leq \delta_0 \text{ then } \pi^* \text{ is the only BO policy for } \hatMDP.
    \end{equation*}
\end{lemma}

\begin{proof}
    Recall Equation (\ref{eq:charact_BO}), the following holds:
    \begin{equation*}
        \pi^* \text{ is BO in } \hatMDP \Longleftrightarrow \hat{\alpha}_s^{\pi^*} \geq 0 \text{ for all } s \in \pi^* \text{ and } \hat{\alpha}_s^{\pi^*} \leq 0 \text{ for all } s \in \calS \setminus \pi^*.
    \end{equation*}

    Let $s \in \calS$ ; because $\pi^*$ is the only BO policy in $\MDP$, by \cite[Lemma 2.(ii)]{gast2023testing} all inequalities are strict:
    \begin{equation*}
        \alpha_s^{\pi^*} > 0 \text{ for all } s \in \pi^* \text{ and } \alpha_s^{\pi^*} < 0 \text{ for all } s \in \calS \setminus \pi^*,
    \end{equation*}
    and by Lemma \ref{lem:conv_of_adv}: 
    \begin{equation*}
    \begin{split}
        \text{for all } s \in \pi^*, \exists \delta^s \text{ s.t. } \norminf{\MDP - \hatMDP} \leq \delta^s \implies \hat{\alpha}_s^{\pi^*} > 0, \\
        \text{for all } s \in \calS \setminus \pi^*, \exists \delta^s \text{ s.t. } \norminf{\MDP - \hatMDP} \leq \delta^s \implies \hat{\alpha}_s^{\pi^*} < 0.        
    \end{split}
    \end{equation*}
    The number of states being finite, defining $\delta_0 := \min_{s \in \calS} \delta^s$ the following holds:
    \begin{equation*}
        \norminf{\MDP - \hatMDP} \leq \delta_0 \implies \hat{\alpha}_s^{\pi^*} > 0 \text{ for all } s \in \pi^* \text{ and } \hat{\alpha}_s^{\pi^*} < 0 \text{ for all } s \in \calS \setminus \pi^*,
    \end{equation*}
    hence the conclusion.
\end{proof}

\subsection{Lemma \ref{lem:bound_slope_adv}: upper bound on the slope of the activation advantage function}

The two previous lemmas provided us with tools for upper bounding the optimal activation advantage gap for all penalties that make the BO policy unique in $\MDP(\lambda)$. For other penalties, we need to bound the slopes of the piecewise affine functions $(\alpha^*_s)_s$ defined in Equation (\ref{eq:act_adv_optimal}).

\begin{lemma}
    \label{lem:bound_slope_adv}
    Let $\pi \in \calA^\calS$ a policy and $s \in \calS$ a state, and suppose that $\MDP$ is unichain. Consider $\alpha_s^\pi (\lambda)$ the activation advantage of state $s$ under policy $\pi$ for the $\lambda-$penalized MDP $\MDP (\lambda)$. It is affine in $\lambda$ and its slope is bounded, in absolute value, by $2 \diam{P^\pi} + 1$.
\end{lemma}

\begin{proof}
    Let $\roundMDP := \big( \calS, \calA, (P^a)_{a \in \calA}, (\mathring{r}^a)_{a \in \calA} \big)$ be the MDP obtained by replacing rewards of $\MDP$ by $1$ for action $1$ and $0$ for action $0$: for every state $s \in \calS, \mathring{r}^1_s := 1$ and $\mathring{r}^0_s := 0$
    Denote by $b^\pi, \mathring{b}^\pi$ and $\alpha^\pi, \mathring{\alpha}^\pi$ the biases and activation advantages vectors under policy $\pi$, in $\MDP$ and $\roundMDP$, respectively.

    Then the reward vector under policy $\pi$ of the $\lambda-$penalized MDP $\MDP (\lambda)$ can be rewritten: $r^\pi(\lambda) = r^\pi (0) - \lambda \mathring{r}^\pi$, and as explained in \cite[Section 3.4, Equation (9)]{gast2023testing}, the bias vector $b^\pi(\lambda)$ is linear in the reward vector $r^\pi(\lambda)$, so that $\forall s' \in \calS$:    
    \begin{equation*}
        b_{s'}^\pi(\lambda) = b_{s'}^\pi (0) - \lambda \mathring{b}_{s'}^\pi,
    \end{equation*}
    and subsequently:
    \begin{equation*}
    \begin{split}
        \alpha_s^\pi (\lambda) & = r^1_s - \lambda - r^0_s + (P^1_{s,\cdot} - P^0_{s,\cdot}) \cdot (b^\pi (0) - \lambda \mathring{b}^\pi)
    \end{split}
    \end{equation*}
    so $\alpha_s^\pi$ is affine in $\lambda$, and:
    \begin{equation*}
    \begin{split}
        \bigg| \frac{\partial \alpha_s^\pi}{\partial \lambda} (\lambda) \bigg| & = \big| -1 + (P^1_{s,\cdot} - P^0_{s, \cdot}) \cdot \mathring{b}^\pi \big| \\
        & \leq 1 + \frac{1}{2} \range{\mathring{b}^\pi} \normone{P^1_{s,\cdot} - P^0_{s,\cdot}} \text{by Equation (\ref{eq:first_dev_lemma})} \\
        & \leq 1 + \range{\mathring{b}^\pi}.
    \end{split}
    \end{equation*}
    Finally, by \cite[Lemma D.3]{boone2025asymptoticallyoptimalregretcommunicating} which we can apply because $\MDP$ is unichain, the following holds:
    \begin{equation*}
        \range{\mathring{b}^\pi} \leq 2 \range{\mathring{r}^\pi} \diam{P^\pi} = 2 \diam{P^\pi}
    \end{equation*}
    hence the conclusion.
\end{proof}

\begin{corollary}
    \label{cor:adv_star_slope}
    For every state $s \in \calS$, $\alpha_s^*$ is $(2 D_{\max} + 1) -$Lipschitz continuous with $D_{\max} := \max_{\pi \in \calA^{\calS}} \diam{P^\pi}$.
\end{corollary}

\begin{proof}
    $\alpha_s^*$ is continuous, affine by parts and by Lemma \ref{lem:bound_slope_adv} on each affine part its slope is bounded by $2 D_{\max} + 1$.
\end{proof}

\subsection{Lemma \ref{lem:technical_cv_fct}: piecewise functions analysis}
\label{subsec:proof_lem_technical_cv}

As a last step to allow the proof of Lemma \ref{lem:proof_th_lin_conv} to put together all previous lemmas, we need a last result related to function analysis.

\begin{lemma}
    \label{lem:technical_cv_fct}
    For some $a < b \in \mathbb{R}$, let $f : [a, b] \rightarrow \mathbb{R}$ be a continuous function that is piecewise affine with a  finite number of pieces. Suppose that there is a unique $\lambda_0 \in [a,b]$ s.t. $f(\lambda_0) = 0$. Then, there exists a constant $c_f$ that only depends on $f$ s.t. for each continuous function $\hat{f} : [a, b] \rightarrow \mathbb{R}$:
    \begin{equation*}
        \text{if } \hat{\lambda} \in [a,b] \text{ satisfies } \hat{f} (\hat{\lambda}) = 0 \text{ then } | \lambda_0 - \hat{\lambda} | \leq c_f \max_{\lambda \in [a,b]} |f(\lambda) - \hat{f}(\lambda)|,
    \end{equation*}
    where $max_{\lambda \in [a,b]} |f(\lambda) - \hat{f}(\lambda)|$ is well defined because $[a,b]$ is compact and $f, \hat{f}$ are continuous.
\end{lemma}

\begin{proof}
    We start by proving the lemma when $f$ is affine on $[a,b]$ and then expanding the result to the case when $f$ is piecewise affine on an arbitrary (but finite) number of intervals. Let $\hat{f} : [a, b] \rightarrow \mathbb{R}$ be a continuous function and $\hat{\lambda} \in [a,b]$ satisfying $\hat{f} (\hat{\lambda}) = 0$.

    \emph{Affine case:} there exists $c, d \in \mathbb{R}$ s.t. $\forall \lambda \in [a,b], f(\lambda) = c \lambda + d$. Because $f$ only vanishes once on $[a,b]$, $s$ is nonzero. . On one hand, $|f(\hat{\lambda})| = |f(\hat{\lambda}) - \hat{f} (\hat{\lambda})| \leq \norminf{f - \hat{f}}$. On the other hand, $|f(\hat{\lambda})| = |f(\hat{\lambda}) - f(\lambda_0)| = |c (\hat{\lambda} - \lambda_0)| = |c| |\hat{\lambda} - \lambda_0|$. Combining both, we get $|\hat{\lambda} - \lambda_0| \leq c_f \norminf{f - \hat{f}}$ with $c_f := \frac{1}{|c|}$.

    \emph{Piecewise affine case:} suppose that $f$ is piecewise affine on $k \in \mathbb{N}$ intervals and let $a =: a_0 < a_1 < \dots < a_k := b$ be s.t. $f$ is affine on each $[a_i, a_{i+1}]$ for $0 \leq i < k$. We proved earlier that for each $0 \leq i < k$, if $\lambda_0 \in [a_i, a_{i+1}]$ there is a constant $c_f(i)$ that only depends on $f$ s.t. for each restricted function $\hat{f}|_{[a_i, a_{i+1}]}$ and $\hat{\lambda} \in [a_i, a_{i+1}]$:
    \begin{equation}
        \label{eq:affine_case}
        \hat{f} (\hat{\lambda}) = 0 \implies | \lambda_0 - \hat{\lambda} | \leq c_f (i) \norminf{f - \hat{f}}.
    \end{equation}

    Now, let us introduce $\delta := \frac{1}{2} \min_{0 \leq i < k} \{ |f(a_i)| ; f(a_i) \neq 0 \}$. There are two cases:
    \begin{enumerate}
        \item If $\norminf{f - \hat{f}} \leq \delta$, there exists an interval on which $f$ is affine and that contains both $\lambda_0$ and $\hat{\lambda}$. Applying (\ref{eq:affine_case}) in this interval proves:
        \begin{equation*}
            | \lambda_0 - \hat{\lambda} | \leq \max_{0 \leq i < k} c_f (i) \norminf{f - \hat{f}}.
        \end{equation*}
        \item If $\norminf{f - \hat{f}} > \delta$, then
        \begin{equation*}
            |\lambda_0 - \hat{\lambda}| \leq b - a \leq \frac{b-a}{\delta} \norminf{f - \hat{f}}.
        \end{equation*}
    \end{enumerate}
    
    Hence the result with $c_f := \max \{ c_f(i) ; 0 \leq i < k \} \cup \{ \frac{b-a}{\delta} \}$.
\end{proof}

\subsection{Main proof}

\begin{proof}[Proof of Lemma \ref{lem:proof_th_lin_conv}]

    Our main goal is to upper bound 
    \begin{equation*}
        \sup_{\text{relevant penalty } \lambda} \norminf{\hat{\alpha}^* (\lambda) - \alpha^*(\lambda)}
    \end{equation*}
    in terms of $\norminf{\hatMDP - \MDP}$, in four steps. First, thanks to previous results \cite{gast2023testing} and Lemma \ref{lem:WI_bounded} we restrict ourselves to an interval that contains all Whittle indices of $\MDP$ and a neighborhood for each of them, thus giving meaning to the aforementioned ``relevant penalties''. Then, we use Lemmas \ref{lem:conv_of_adv} and \ref{lem:uniqueBO_close} to state that if some penalty $\lambda$ is not close to any of the Whittle indices of $\MDP(\lambda)$, then the BO policy is unique and the same as in $\hatMDP(\lambda)$, consequently the activation advantage difference can be bounded easily. As a third step, Lemma~\ref{lem:bound_slope_adv} provides tools to bound that difference even for penalties that are close to the Whittle indices of $\MDP$, where BO policies might not be the same in $\MDP(\lambda)$ and $\hatMDP(\lambda)$. Finally, we apply Lemma \ref{lem:technical_cv_fct} to conclude.

    From \cite[Corollary 3]{gast2023testing} all the Whittle indices of $\MDP$ are finite. Let $\nu := \frac{1}{2} \min_{\lambda_s \neq \lambda_{s'}} |\lambda_s - \lambda_{s'}|$ and define a ``left bound'' and a ``right bound'':
    \begin{equation*}
        \begin{split}
            \lambda_l & := \min \{  \min_{s \in \calS} \lambda_s - \nu, 
        - \norminf{r^1 - r^0} - \range{r^1} \diam{P^1} (\norminf{P^1 - P^0} + 2 \nu)\} \\
        \lambda_r & := \max \{ \max_{s \in \calS} \lambda_s + \nu, 
            \norminf{r^1 - r^0} + \range{r^0} \diam{P^0} (\norminf{P^1 - P^0} + 2 \nu)\}.
        \end{split}
    \end{equation*}
    Denote by $I$ the closed interval $[\lambda_l, \lambda_r]$; by Lemma \ref{lem:WI_bounded} it contains all Whittle indices of $\MDP$ and a non-empty neighborhood for each of them. To prove the result our goal is to first bound $\max_{\lambda \in I} \norminf{\alpha^*(\lambda) - \hat{\alpha}^*(\lambda)}$ the optimal activation advantage difference on $I$, and then apply Lemma \ref{lem:technical_cv_fct} to conclude.

    \emph{Bounding the optimal activation advantage difference:} to ease the notation, define $\varepsilon_{\hatMDP} := \norminf{\MDP - \hatMDP}$. Note that Assumption \ref{ass:tildeM_close} implies $\varepsilon_{\hatMDP} \leq \nu$, so that if $s,s' \in \calS$ satisfy $\lambda_s < \lambda_{s'}$ then $\lambda_s + \varepsilon_{\hatMDP} \leq \lambda_{s'} - \varepsilon_{\hatMDP}$.
    
    Let us now consider two successive and distinct Whittle indices $\lambda_s < \lambda_{s'}$ when taken in increasing order (i.e. no other state $s'' \in \calS$ satisfy $\lambda_s < \lambda_{s''} < \lambda_s'$). By Lemma \ref{lem:characterization_WI} there is a unique policy $\pi^*$ that is BO in both penalized MDPs $\MDP(\lambda_s + \varepsilon_{\hatMDP})$ and $\MDP(\lambda_{s'} - \varepsilon_{\hatMDP})$. So by Lemma \ref{lem:uniqueBO_close} there exists thresholds $\varepsilon_{s+} > 0$ and $\varepsilon_{s'-} > 0$ such that:
    \begin{equation*}
        \varepsilon_{\hatMDP} \leq \min \{ \varepsilon_{s+}, \varepsilon_{s'-} \} \implies \pi^* \text{ is the only BO policy in both $\hatMDP(\lambda_s + \varepsilon_{\hatMDP})$ and $\hatMDP(\lambda_{s'} - \varepsilon_{\hatMDP})$.}
    \end{equation*}

    For every penalties $\lambda_0 < \lambda_1$, a policy $\pi$ that is BO in both $\MDP(\lambda_0)$ and $\MDP(\lambda_1)$ satisfies:
    \begin{equation*}
    \begin{split}
        \forall \lambda \in \{\lambda_0, \lambda_1\}, \alpha^{\pi}_s(\lambda) \geq 0 & \text{ for every } s \in \pi \\
        \forall \lambda \in \{\lambda_0, \lambda_1\}, \alpha^{\pi}_s(\lambda) \leq 0 & \text{ for every } s \in \calS \setminus \pi.
    \end{split}
    \end{equation*} 
    As the activation advantage function $\alpha_s^\pi(\lambda)$ is affine in $\lambda$, these two facts remain true for every $\lambda \in [\lambda_0, \lambda_1]$ and $\pi$ is BO for every MDP $\MDP(\lambda)$ with $\lambda \in [ \lambda_0, \lambda_1 ]$. Similarly, a policy that is not BO in both $\MDP(\lambda_0)$ and $\MDP(\lambda_1)$ remains non-BO for every MDP $\MDP(\lambda)$ with $\lambda \in [ \lambda_0, \lambda_1 ]$. As a consequence:
    \begin{equation*}
        \varepsilon_{\hatMDP} \leq \min \{ \varepsilon_{s+}, \varepsilon_{s'-} \} \implies \pi^* \text{ is the only BO policy in $\hatMDP(\lambda)$ for every $\lambda \in [\lambda_s + \varepsilon_{\hatMDP}, \lambda_{s'} - \varepsilon_{\hatMDP}]$},
    \end{equation*}
    which proves Lemma \ref{lem:proof_th_lin_conv}\ref{lem:proof_th_lin_conv:uniqueBO} with $\delta_{\MDP} := \min \bigcup_{s \in \calS} \{ \varepsilon_s^-, \varepsilon_s^+ \}$. Moving on to proving Lemma \ref{lem:proof_th_lin_conv}\ref{lem:proof_th_lin_conv:zero}, the optimal activation advantages of $\MDP$ and $\hatMDP$ can be rewritten:
    \begin{equation*}
        \forall \lambda \in [\lambda_s + \varepsilon_{\hatMDP}, \lambda_{s'} - \varepsilon_{\hatMDP}], \alpha^*(\lambda) = \alpha^{\pi^*} (\lambda) \text{ and } \hat{\alpha}^*(\lambda) = \hat{\alpha}^{\pi^*} (\lambda),
    \end{equation*}
    and Corollary \ref{cor:WI_unif_cv_of_adv} can be applied, assuming $\varepsilon_{\hatMDP} \leq \min_{s \in \calS} \{ \varepsilon_{s,+}, \varepsilon_{s,-} \}$:
    \begin{equation}
        \label{eq:adv_bound_in_J}
        \exists c_\lambda > 0 \text{ that only depends on $\MDP, \lambda_l$ and $\lambda_r$ s.t. } \max_{\lambda \in J} \norminf{\alpha^*(\lambda) - \hat{\alpha}^*(\lambda)} \leq c_\lambda \norminf{\MDP - \hatMDP},
    \end{equation}
    where $J := I \setminus \bigcup_{s \in \calS} [\lambda_s - \varepsilon_{\hatMDP}, \lambda_{s} + \varepsilon_{\hatMDP}]$ denotes the penalties of $I$ that are at distance greater than $\varepsilon_{\hatMDP}$ from the Whittle indices of $\MDP$. Note that $\lambda_l$ and $\lambda_r$ only depend on $\MDP$ therefore $c_\lambda$ only depends on $\MDP$.

    We prove similarly as before that:
    \begin{equation*}
        \forall \lambda \in J, \norminf{\alpha^* (\lambda) - \hat{\alpha}^* (\lambda)} \leq 8 (r_{\max} + |\lambda|) \max_{\pi \in \calA^{\calS}} \diam{P^{\pi}} \norminf{\MDP - \hatMDP}
    \end{equation*}
    which does only depend on $\MDP$, therefore Equation (\ref{eq:adv_bound_in_J}) also holds assuming $\varepsilon_{\hatMDP} > \min_{s \in \calS} \{ \varepsilon_{s,+}, \varepsilon_{s,-} \}$ with $c_\lambda := 8 (r_{\max} + \max \{ |\lambda_l|, |\lambda_r|\}) \max_{\pi \in \calA^{\calS}} \diam{P^{\pi}} / \min_{s \in \calS} \{ \varepsilon_{s,+}, \varepsilon_{s,-} \}$.
    
    What is left for this part is to also bound $\norminf{\alpha^*(\lambda) - \hat{\alpha}^*(\lambda)}$ around the Whittle indices of $\MDP$, that is for $\lambda \in I \setminus J$. By Corollary \ref{cor:adv_star_slope}, for every state $s\in \calS$ the optimal activation advantage function $\alpha_s^*$ is $(2 D_{\max} + 1) -$Lipschitz continuous with $D_{\max} := \max_{\pi \in \calA^{\calS}} \diam{P^\pi}$, and $\hat{\alpha}_s^*$ is $(2 \hat{D}_{\max} + 1) -$Lipschitz continuous with $\hat{D}_{\max} := \max_{\pi \in \calA^{\calS}} \diam{\hat{P}^\pi}$. By \cite[Equation D.12]{boone2025asymptoticallyoptimalregretcommunicating} and Assumption \ref{ass:tildeM_close}, $\hat{D}_{\max} \leq 2 D_{\max}$ so, all in all:
    \begin{equation*}
        \hat{\alpha}_s^* \text{ is } (4 D_{\max} + 1) -\text{Lipschitz continuous, and } \alpha_s^* \text{ is } (2 D_{\max} + 1) -\text{Lipschitz continuous}.
    \end{equation*}
    Consequently, let us take some $\lambda \in \bigcup_{s \in \calS} [\lambda_s, \lambda_s + \varepsilon_{\hatMDP}]$. There exists a state $s' \in \calS$ s.t. $\lambda \in [\lambda_{s'}, \lambda_{s'} + \varepsilon_{\hatMDP}]$, so for every $s \in \calS$:
    \begin{equation*}
    \begin{split}
        | \alpha_s^* (\lambda) - \hat{\alpha}_s^* (\lambda) | & = | \alpha_s^* (\lambda) - \alpha_s^*(\lambda_{s'} + \varepsilon_{\hatMDP}) + \alpha_s^*(\lambda_{s'} + \varepsilon_{\hatMDP}) - \hat{\alpha}_s^* (\lambda) | \\
        & \leq (2 D_{\max} + 1) \varepsilon_{\hatMDP} + |\alpha_s^*(\lambda_{s'} + \varepsilon_{\hatMDP}) - \hat{\alpha}^*_s (\lambda)| \text{ by Lipschitz continuity of } \alpha_s^* \\
        & \leq (6 D_{\max} + 2) \varepsilon_{\hatMDP} + |\alpha_s^*(\lambda_{s'} + \varepsilon_{\hatMDP}) - \hat{\alpha}^*_s (\lambda_{s'} + \varepsilon_{\hatMDP})| \text{ by Lipschitz continuity of } \hat{\alpha}_s^* \\
        & \leq (c_\lambda + 6 D_{\max} + 2) \varepsilon_{\hatMDP} \text{ by Equation (\ref{eq:adv_bound_in_J})},
    \end{split}
    \end{equation*}
    and the same bound holds when $\lambda \in \bigcup_{s \in \calS} [\lambda_s - \varepsilon_{\hatMDP}, \lambda_s]$ with a completely analogous proof. Combining these two bounds and the one of Equation (\ref{eq:adv_bound_in_J}):
    \begin{equation*}
        \max_{\lambda \in I} \norminf{\alpha^*(\lambda) - \hat{\alpha}^*(\lambda)} \leq (c_\lambda + 6 D_{\max} + 2) \norminf{\MDP - \hatMDP}.
    \end{equation*}

    \emph{Concluding the proof.} Let now $s \in \calS$ be a state and let us prove the existence of some constant $c_\MDP$ that only depends on $\MDP$ such that every $\hat{\lambda} \in I$ that satisfies $\hat{\alpha}_s^*(\hat{\lambda}) = 0$ also satisfies:
    \begin{equation*}
        |\lambda_s - \hat{\lambda}| \leq c_\MDP \norminf{\MDP - \hatMDP}
    \end{equation*}
    Recall, $\alpha_s^*(\lambda)$ is continuous and piecewise affine in $\lambda$. Because $\MDP$ is indexable, there is a unique $\lambda_s \in I$ satisfying $\alpha_s^*(\lambda_s) = 0$.

    In addition, by Lemma \ref{lem:WI_bounded} the following holds:
    \begin{itemize}
        \item $\lambda \leq - \norminf{r^1 - r^0} - \frac{1}{2} \range{r^1} \diam{\hat{P}^1} \norminf{\hat{P}^1 - \hat{P}^0} \implies \calS$ is BO in $\MDP$,
        \item $\lambda \geq \norminf{r^1 - r^0} \frac{1}{2} \range{r^0} \diam{\hat{P}^0} \norminf{\hat{P}^1 - \hat{P}^0} \implies \emptyset$ is BO in $\MDP$.
    \end{itemize}
    From \cite[Equation D.12]{boone2025asymptoticallyoptimalregretcommunicating} and Assumption \ref{ass:tildeM_close}, $\diam{\hat{P}^a} \leq 2 \diam{P^a}$ for every $a \in \calA$. Moreover:
    \begin{equation*}
        \norminf{\hat{P}^1 - \hat{P}^0} \leq \norminf{P^1 - P^0} + 2 \varepsilon_{\hatMDP} \leq \norminf{P^1 - P^0} + 2 \nu.
    \end{equation*}
    Combining these facts yields:
    \begin{itemize}
        \item $\lambda \leq - \norminf{r^1 - r^0} - \range{r^1} \diam{P^1} (\norminf{P^1 - P^0} + 2 \nu) \implies \calS$ is BO in $\hatMDP$,
        \item $\lambda \geq \norminf{r^1 - r^0} \range{r^0} \diam{P^0} (\norminf{P^1 - P^0} + 2 \nu) \implies \emptyset$ is BO in $\hatMDP$.
    \end{itemize}     
    As a result, $\hat{\alpha}_s^* (\lambda_l) \geq 0$ and $\hat{\alpha}_s^* (\lambda_r) \leq 0$, so, $\alpha^*_s$ being continuous there is at least one $\hat{\lambda} \in [ \lambda_l, \lambda_r ]$ s.t. $\hat{\alpha}^*_s(\hat{\lambda}) = 0$. Applying Lemma \ref{lem:technical_cv_fct}, there is a constant $c_{s}$ that only depends on $\alpha^*_s$ (so only on $\MDP$) s.t.
    \begin{equation*}
        \begin{split}
        | \lambda_s - \hat{\lambda} | & \leq c_{s} \max_{\lambda \in I}{| \alpha^*(\lambda) - \hat{\alpha}^*(\lambda) |} \\
        & \leq c_{s} (c_\lambda + 6 D_{\max} + 2) \norminf{\MDP - \hatMDP}.
        \end{split}
    \end{equation*}
    wich concludes the proof with $c_\MDP := (\max_{s \in \calS} c_s) ( c_\lambda + 6 D_{\max} + 2 )$.
\end{proof}